\documentclass[sigconf]{acmart}
\usepackage{multirow}
\usepackage{pifont}
\usepackage{xcolor}
\usepackage{colortbl}
\usepackage{array}
\usepackage{makecell}
\usepackage{appendix}
\newcolumntype{P}[1]{>{\centering\arraybackslash}p{#1}}
\AtBeginDocument{%
  }

\copyrightyear{2026}
\acmYear{2026}
\setcopyright{cc}
\setcctype{by}
\acmConference[MM '26]{Proceedings of the 34th ACM International Conference on Multimedia}{November 10--14, 2026}{Rio de Janeiro, Brazil}
\acmBooktitle{Proceedings of the 34th ACM International Conference on Multimedia (MM '26), November 10--14, 2026, Rio de Janeiro, Brazil}
\acmDOI{10.1145/3767308.3835229}
\acmISBN{979-8-4007-2213-4/2026/11}

\begin{document}

\title{Unlocking Motion in Expressions: Temporal Calibration for Referring Video Object Segmentation}

\author{Yiwen Jiang}
\affiliation{%
  \institution{School of Computer Science \& Technology}
  \institution{Soochow University}
  \city{Suzhou}
  \country{China}
}
\email{ywjiang7@stu.suda.edu.cn}

\author{Zhengtong Zhu}
\affiliation{%
  \institution{School of Computer Science \& Technology}
  \institution{Soochow University}
  \city{Suzhou}
  \country{China}
}
\email{20245227002@stu.suda.edu.cn}

\author{Ruixin Zhang}
\affiliation{%
  \institution{School of Computer Science \& Technology}
  \institution{Soochow University}
  \city{Suzhou}
  \country{China}
}
\email{20245227041@stu.suda.edu.cn}

\author{Jiaqing Fan}
\authornote{Corresponding author}
\affiliation{%
  \institution{School of Computer Science \& Technology}
  \institution{Soochow University}
  \city{Suzhou}
  \country{China}
}
\email{jqfan@suda.edu.cn}

\renewcommand{\shortauthors}{Yiwen Jiang, Zhengtong Zhu. Ruixin Zhang, and Jiaqing Fan}
\newcommand{\cmark}{\ding{51}}  
\newcommand{\xmark}{\ding{55}}  

\begin{abstract}
 Referring Video Object Segmentation (RVOS) aims to segment referred objects at the pixel level in video sequences based on natural language descriptions. Existing methods typically introduce motion information within a unified cross-modal temporal modeling framework, where language cues are used for target localization and segmentation. However, the dependency of expressions on motion semantics is not explicitly modeled, making it difficult to adaptively adjust the use of motion information according to different semantic requirements. To address these issues, we propose an Expression-driven Motion Calibration (EMC) framework for RVOS that explicitly unlocks and leverages the motion semantics within expressions. The proposed method extracts interpretable motion control signals from expressions via a Motion Signal Processing (MSP) module, and employs a Motion Influence Calibration (MIC) module to adjust the contribution of motion cues during temporal decision making. In addition, a Semantic Temporal Stage Construction (STSC) module is introduced to build expression-relevant temporal stages, providing a compact temporal candidate space for motion calibration. Through extensive evaluation on six standard benchmarks, including Ref-YouTubeVOS, Ref-DAVIS17,
MeViS (valid/valid$^u$), A2D-Sentences, and JHMDB-Sentences, the superiority of our method is validated. We will release the code on \href{https://github.com/Jeven7/EMC}{\textcolor{magenta}{https://github.com/Jeven7/EMC}}.
\end{abstract}





\maketitle
\begin{figure}[h]
\includegraphics[width=\columnwidth]{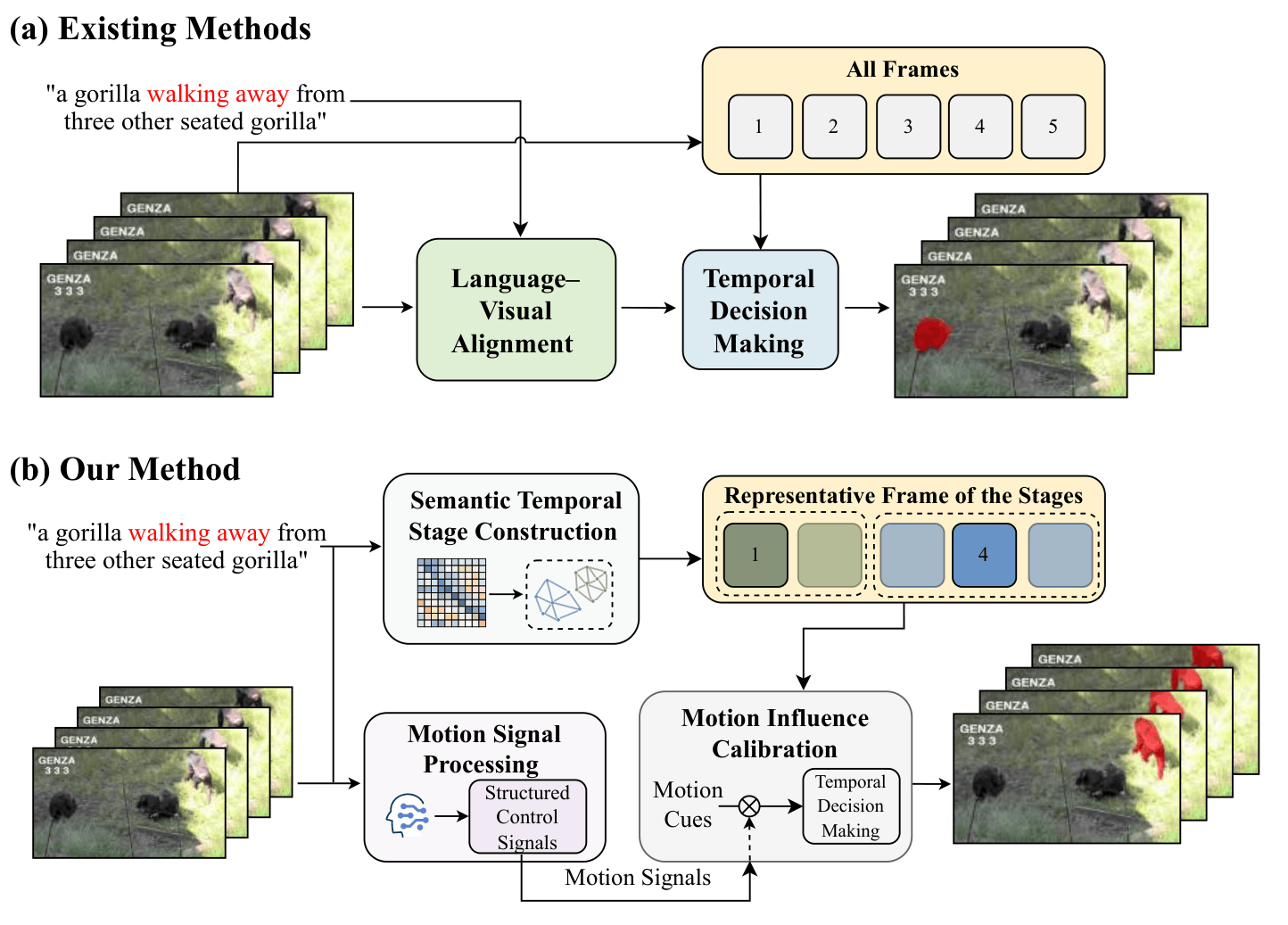} 
\caption{Overview of EMC. (a) Existing RVOS methods typically apply motion modeling to entire video sequences, incorporating motion cues into expressions without explicit differentiation. (b) Our method models expression-related motion importance as motion signals and calibrates motion effects within the local structured semantic temporal space. } 
\label{Fig.main1} 
\end{figure}

\section{Introduction}
Referring Video Object Segmentation (RVOS) aims to perform pixel-level segmentation of target objects within video sequences based on language descriptions\cite{wu2022language,liang2025referdino,videoseg}. Unlike semi-supervised video object segmentation, which relies on pixel-level annotations in the first frame\cite{cheng2024putting,Baek_2025_ICCV,wang2025flowtrack}, or unsupervised video object segmentation, which identifies salient motion regions without explicit object guidance\cite{Huang_2025_CVPR,zheng2025shallow,fan2025periodvos}, RVOS uses language as the means of target specification, requiring models to jointly possess accurate visual understanding and fine-grained semantic parsing capabilities in complex video scenarios. Meanwhile, recent advances in multimodal learning have promoted the alignment between visual and linguistic information, ranging from vision-language perception and multimodal reasoning to large-scale multimodal foundation models\cite{qiao2024dntextspotter,gao2025interleaved,gao2025aim}. Owing to its practical value in applications such as video editing, human–computer interaction, and intelligent content analysis, this task has attracted increasing research attention in recent years. However, due to the high diversity and abstract nature of natural language expressions, together with the presence of complex object motions, occlusions, and appearance variations in videos, effectively aligning language semantics with temporal video information remains a core challenge in RVOS.

In recent years, with the introduction of Mevis\cite{ding2023mevis}, RVOS methods have gradually shifted their research focus toward motion expressions, attempting to further enhance model segmentation capabilities in complex motion scenarios. Existing RVOS approaches generally treat motion as an inherent temporal property of video data and incorporate it through unified temporal representations that encode object evolution across frames\cite{botach2022end,wu2022language,wu2023segment,luo2023soc,tang2023temporal}. These methods incorporate language as a global semantic condition for target localization and segmentation. Recent work has begun to focus more directly on motion information itself, either through explicit modeling of motion-related language descriptions or by decoupling static appearance cues from motion cues in modeling\cite{yuan2024losh,tang2023temporal}, to enhance segmentation performance in scenarios involving complex motion expressions. However, most of these methods generally uniformly employ motion cues without distinguishingly adjusting the level of involvement in motion information based on the importance of motion semantics in expressions. This limits the models' adaptability to the semantic demands of different expressions.

To address these challenges, We propose EMC, an expression-driven motion calibration framework(Figure~\ref{Fig.main1}) that modulates the influence of motion cues in temporal decision making by explicitly modeling the importance of motion semantics. Specifically, we design a Motion Signal Processing (MSP) module that analyzes motion-related semantic components in expressions and explicitly models the importance of motion semantics as an interpretable temporal control prior. MSP decomposes motion importance into two complementary aspects: whether motion cues are necessary under the given expression, and when necessary, how strongly they should influence temporal decision making process. Based on this expression-driven prior, we further introduce a Motion Influence Calibration (MIC) module, which adaptively calibrates the influence of motion cues during temporal decision making process, suppressing redundant motion noise in static expressions while enhancing discriminative motion cues in dynamic ones.

Additionally, when introducing expression-driven motion calibration mechanisms into RVOS, we face the challenge of determining an appropriate temporal influence space within video sequences. Directly adjusting motion influence at the per-frame level is computationally redundant and susceptible to noise frame interference. To address this, we propose the Semantic Temporal Stage Construction (STSC) module. By organizing videos into semantic temporal stages through text-guided temporal spectrum clustering, STSC reduces redundancy and improves efficiency while ensuring temporal coverage. This approach provides compact, expression-relevant candidate temporal structures for subsequent motion influence calibration. Together with MSP and MIC, STSC enables expression-aware motion calibration by providing a structured temporal space for adaptive motion reasoning.
Finally, to comprehensively validate its effectiveness, we evaluated our method on six RVOS benchmarks: Ref-Youtube-VOS\cite{seo2020urvos}, MeViS(valid/valid$^u$)\cite{ding2023mevis}, Ref-DAVIS17\cite{davis}, A2D-Sentences\cite{gavrilyuk2018actor} and JHMDB-Sentences\cite{gavrilyuk2018actor}. Experimental results show that our approach achieves state-of-the-art performance across multiple metrics on diverse benchmarks. Overall, our contributions are summarized as follows:
\begin{itemize}
    \item We propose EMC, a novel expression-driven motion calibration framework for RVOS. This framework extracts motion significance from expressions as control signals through a motion signal processing (MSP) module, guiding more accurate motion-aware temporal decisions making process.
    \item We introduce a Motion Influence Calibration (MIC) module, which adaptively regulates the relative importance and contribution of motion cues during temporal decision making utilizing the motion signals explicitly extracted from MSP.
    \item We design a Semantic Temporal Stage Construction (STSC) module, which organizes videos into expression-related temporal stages, thereby providing a concise and compact temporal structure for subsequent motion calibration.
    \item Our method achieves competitive result across six datasets, with consistent improvements over existing methods.
\end{itemize}
\section{Related Work}
\subsection{Referring Video Object Segmentation}
 RVOS aims to segment target objects in videos based on language descriptions\cite{mutr,vdit,pan2025semantic}. Early RVOS approaches primarily extended image referring segmentation to video scenarios\cite{seo2020urvos}. With the introduction of DETR-based architectures, the query-based end-to-end paradigm became dominant in RVOS research\cite{botach2022end,wu2022language,luo2023soc}, Language were explicitly incorporated as object queries to guide the spatio-temporal decoding process within unified models. This paradigm enables direct cross-modal alignment and simplifies the optimization pipeline. Over the past three years, RVOS research has shifted toward scalability, robustness, and deployment requirements in complex scenarios. A series of works\cite{wu2023onlinerefer,han2023html} explored online or long video settings, enhancing temporal consistency through mechanisms like query propagation or representation updates. These methods emphasize efficient temporal modeling under streaming or resource-constrained conditions. Concurrently, with the emergence of large scale benchmarks like MeViS\cite{ding2023mevis}, RVOS has been systematically studied under more challenging expressions and scenarios, further integrating with foundational models and large scale visual-language systems\cite{liang2025referdino,cuttano2025samwise,wei2025instructseg,yan2024visa,MPG-SAM2}. Such integration significantly improves generalization ability and semantic understanding across diverse visual contexts, benefiting from multimodal foundation models and their applications in visual understanding and video content creation \cite{gao2025aim,zhang2024musetalk,zhong2025anytalker}.Unlike highly coupled end-to-end paradigms, Findtrack\cite{findtrack} proposes decoupling object recognition from temporal propagation. This approach demonstrates stronger robustness in referential ambiguity and complex scenarios, highlighting the potential advantages of structured modeling in RVOS.
\subsection{Motion Expression Video Segmentation}
Compared to traditional RVOS tasks, this task primarily segments objects based on their motion descriptions within the video. This shift places greater emphasis on capturing fine-grained temporal dynamics and motion-aware semantics. Early LMPM\cite{ding2022language} explored the interaction between linguistic information and temporal visual features, enhancing cross-frame object understanding through language-guided spatio-temporal modeling. However, motion semantics remained unified with appearance information in the model. Subsequently, DsHmp\cite{DsHmp} explicitly distinguished static perception from hierarchical motion perception, introducing a motion perception mechanism to better differentiate objects with similar appearances but distinct motion patterns. LoSh\cite{yuan2024losh}, approaching from the linguistic side, noted that long term representations often contain motion or temporal descriptions. It effectively mitigated the model's over reliance on motion semantics by jointly modeling long term and short term expressions. DMVS\cite{fang2025decoupled} further decoupled the understanding of motion expressions from video instance segmentation, modeling static and motion cues separately. This significantly improved segmentation robustness in expression scenarios dominated by motion semantics. Although the above methods incorporate motion expression modeling at different levels, most still employ motion cues in a uniform manner. They fail to differentiate modeling based on the semantic importance of motion within expressions, potentially resulting in limited adaptability in expression scenarios with varied semantic requirements in practice, highlighting the need for more adaptive modeling strategies.
\begin{figure*}[t]
    \begin{center}

   \includegraphics[width=0.98\linewidth]{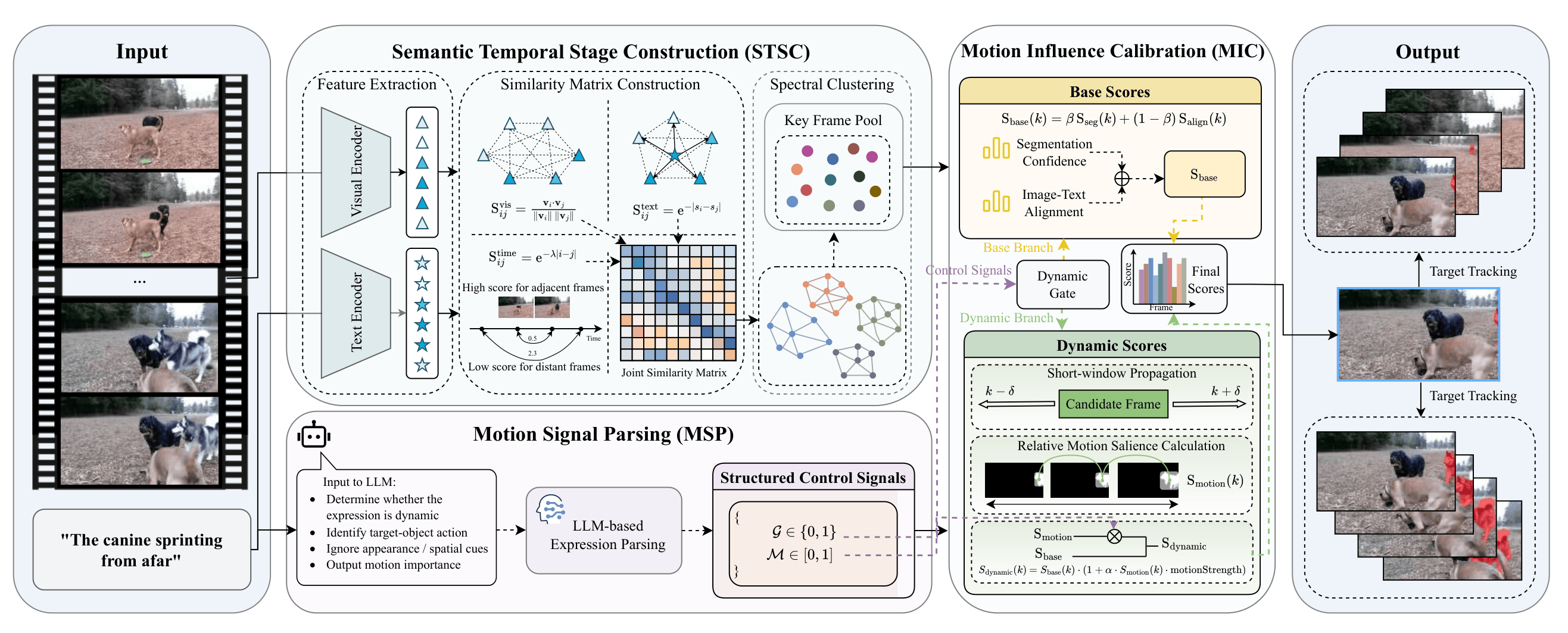}
\end{center}
\vspace{-0.5em}
    \caption{The architecture of EMC. First, STSC constructs expression-related temporal phases based on input video and expressions, defining a compact and efficient temporal influence space for subsequent adaptive temporal decision making process. Simultaneously, MSP models the importance of motion semantics within gesture expressions to extract explicit motion control signals. Finally, MIC calibrates the contribution of motion cues during the temporal decision making process. Through this approach, motion information can be utilized selectively and explicitly according to the semantic requirements of different expressions, thereby more effectively enhancing the robustness and flexibility of the RVOS system in complex scenarios.}
    \label{Fig.main2}
\end{figure*}
\section{Methodology}
\subsection{Overview}
The overall architecture of our proposed method is illustrated in Figure~\ref{Fig.main2}, primarily consisting of three core modules. Given an input video $\mathcal{V} = \{ I_t \}_{t=1}^{T}$ composed of $T$ frames and a referring expression $E$, the goal of referring video object segmentation is to predict a sequence of segmentation masks $\{ \hat{Y_t} \}_{t=1}^{T}$ that localize the target object specified by in each frame. First, STSC comprehensively models the input video and expression across three levels, constructing expression-related semantic temporal stages. It extracts candidate key frames for subsequent decision making in motion calibration. Simultaneously, MSP leverages LLM to semantically analyze input expressions, assess the significance of motion cues within them, and extract explicit motion control signals, which are then passed to MIC. Subsequently, MIC applies differentiated processing to static and dynamic expressions based on these control signals. By adjusting the participation level of motion saliency in the keyframe selection process, it calibrates the keyframe decisions for dynamic expressions. Finally, based on the segmentation results of the selected key frames, the model employs a bi-directional propagation strategy to generate a complete video segmentation mask.
\subsection{Semantic Temporal Stage Construction}
STSC models the semantic temporal structure of a video by jointly considering visual consistency, text-guided semantic relevance, and temporal proximity. By constructing a joint similarity matrix and applying spectral clustering, the video is partitioned into a set of semantic temporal stages, within which frames are both visually coherent and consistently aligned with the referring expression. Representative frames with the highest text relevance from each stage are then selected into a keyframe pool for subsequent process.

\textbf{Text-Guided Consistency Similarity.}
We introduce a text-guided consistency similarity to ensure that frames within the semantic temporal stage exhibit consistent relevance to the referring expression. Specifically, we compute an expression relevance score $s_i$ for each frame based on semantic correspondence with expression and define the text-guided consistency similarity as:
\begin{equation}
S^{\text{text}}_{ij}=\exp(-|s_i-s_j|),
\end{equation}
where $s_i$ denotes the expression relevance score of frame $i$. This formulation encourages frames with similar semantic responses to be grouped into the same temporal stage while suppressing semantically inconsistent frames during clustering.

\textbf{Temporal Proximity Similarity.}
To encourage temporal continuity within semantic temporal stages, we introduce a temporal proximity similarity $S^{\text{time}}$ to model the temporal distance between frames. This similarity provides a local temporal smoothness constraint during stage construction, encouraging temporally adjacent frames to be grouped together while avoiding excessive connections between distant frames. Formally, given the timestamps $t_i$ and $t_j$ of frames $i$ and $j$, the similarity is defined as:
\begin{equation}
S_{ij}^{\text{time}} = e^{-\lambda \lvert t_i - t_j \rvert},
\end{equation}
where $\lambda$ is a decay factor controlling the impact of temporal distance. This formulation assigns higher similarity to temporally adjacent frames while reducing the influence of distant frame pairs, thereby preserving temporal coherence within semantic temporal stages.

\textbf{Visual Similarity.} 
We introduce a visual similarity $S^{\text{vis}}$ to measure appearance consistency between frames. Given corresponding visual features $v_i$ and $v_j$ extracted from frames $i$ and $j$, respectively, the visual similarity is defined as the cosine similarity:
\begin{equation}
S^{\text{vis}}_{ij}=\frac{v_i^\top v_j}{\|v_i\|\|v_j\|},
\end{equation}
which provides a normalized measure of feature alignment and reduces the influence of feature scale variations. This similarity encourages visually consistent frames to be grouped into the same semantic temporal stage, providing a stable appearance constraint for reliable stage construction and improving robustness against transient motion and background noise in complex scenarios.

\textbf{Temporal Stage Aggregation.}
Based on the three similarity terms, we jointly aggregate them to construct a joint similarity matrix that captures visual coherence, expression relevance consistency, and temporal proximity between frames.We first normalize the matrices of the three categories to obtain normalized similarity values $\widetilde{S}^{\text{text}}$, $\widetilde{S}^{\text{time}}$, and $\widetilde{S}^{\text{vis}}$, ensuring comparability among different complementary similarity cues during fusion. We then compute the overall similarity between frames $i$ and $j$ as a balanced weighted combination of the normalized similarities:
\begin{equation}
    S_{ij} =
    \frac{
    w_t \, \widetilde{S}_{ij}^{\text{text}} +
    w_\tau \, \widetilde{S}_{ij}^{\text{time}} +
    w_v \, \widetilde{S}_{ij}^{\text{vis}} 
    }{
    w_t + w_\tau + w_v
    },
\end{equation}%
where $w_t$, $w_\tau$ and $w_v$ denote the weights for text-guided,  temporal, and visual similarities, respectively. Based on the resulting joint similarity matrix $S$, we apply spectral clustering to partition the video into $n_{cluster}$ semantic temporal stages.

For each temporal stage, representative key frames are selected according to their expression relevance scores and then collected into a keyframe pool $\mathcal{K}$, which serves as the structured temporal candidate frame set for subsequent motion-aware calibration.

\subsection{Motion Signal Processing}
MSP models the importance of motion semantics in the referring expression at the expression level and represents it as a structured semantic signal, which characterizes the extent to which the current expression relies on motion cues for target grounding. Rather than modeling the specific content of motion semantics, MSP focuses primarily on the relative weight of motion cues within the overall expression semantics. Specifically, MSP captures motion importance from two complementary perspectives: it determines whether motion cues are required under the given expression and, if so, quantifies how strongly they should influence the subsequent temporal decision-making process.
Unlike approaches that implicitly encode motion-related information into high-dimensional language representations, MSP explicitly formulates the latent motion importance in the expression as a low-dimensional and interpretable semantic prior. This design enables the dependency on motion cues from the language side to be more clearly separated and modeled.

MSP operates purely on the linguistic modality and leverages LLM to perform semantic analysis of the referring expression. Through language-level parsing, the model determines whether the expression involves explicit motion semantics associated with the target object and estimates their relative importance under the given expression. The output of MSP is defined as a structured and interpretable motion control signal, denoted as $\mathrm{MSP}(E) = \{\mathcal{G},\ \mathcal{M}\}$, 
where $\mathcal{G}\in\{0,1\}$ indicates whether the expression contains motion semantics, and $\mathcal{M}\in[0,1]$ quantifies the importance of motion semantics within the overall expression.

The resulting motion strength serves as an expression-conditioned semantic prior and is used as a gating factor in the subsequent MIC process, indicating the extent to which motion cues should participate in the scoring function. It does not represent the physical magnitude of motion in the video, nor does it directly produce motion saliency; instead, it provides a language-driven quantification of motion importance, thereby modulating the contribution of the motion term during key frame scoring. To ensure semantic consistency, MSP focuses exclusively on motion performed by the referred object itself and ignores non-action factors such as appearance attributes and spatial descriptions; when no motion semantics are present in the expression, $\mathcal{M}$ is set to zero.

By explicitly modeling motion semantic importance as a structured low-dimensional signal, MSP effectively establishes a clear interface between language understanding and temporal decision making, providing an expression-driven semantic guidance prior for the subsequent adaptive motion calibration process.

\subsection{Motion Influence Calibration}
MIC module incorporates motion cues into the key frame scoring process as calibration factors under expression conditions, rather than directly involving them in decision making as a unified temporal signal. MIC first calculates a base reliability score for each candidate frame based on $E$, which serves as the primary criterion for key frame selection. Subsequently, MIC further calibrates this base score using local motion salience near the candidate frame as an auxiliary cue, with the calibration weight controlled by the motion importance signal output from MSP. MIC outputs key frame scores based on the candidate key frame set constructed by STSC under expression conditions, which are used for subsequent key frame selection and temporal propagation.

\textbf{Base Reliability Scoring.}
Upon receiving the motion signal processed by the MPS, the MIC first calculates the base reliability score for candidate frames. The base reliability score for candidate frame $k$ measures whether the frame can independently support target referencing under a given expression without relying on motion cue calibration. This score is composed by the prediction confidence of the target from single frame segmentation results and the semantic alignment between the predicted mask region and the expression. The base reliability score is calculated as follows:
\begin{equation}
    S_{\mathrm{base}}(k)
    =
    \beta \, S_{\mathrm{seg}}(k)
    +
    (1-\beta)\, S_{\mathrm{align}}(k),
\end{equation}%
where $S_{\mathrm{seg}}(k)$ denotes the segmentation confidence of the predicted mask on frame $k$, and $S_{\mathrm{align}}(k)$ measures the semantic consistency between the mask region and the referring expression. The weighting factor $\beta$ balances visual reliability and language alignment. The result will be sent to the temporal decision making part.

\textbf{Motion Calibration.}
Motion Calibration introduces motion cues as an expression-conditioned correction to the base reliability score. For each candidate frame $k$, we first estimate a local motion saliency score $S_{\mathrm{motion}}(k)$ to characterize the relative motion strength of the referred object in a short temporal neighborhood around $k$.Starting from the predicted object mask on frame $k$, the mask is propagated forward and backward within a local temporal window $[k-\delta,\; k+\delta]$ For each propagated frame $t$, we compute the centroid of the object mask and measure the displacement between adjacent frames. The average centroid displacement within the window is used to define the motion saliency score:
\begin{equation}
    S_{\mathrm{motion}}(k)
    =
    \frac{1}{2\delta}
    \sum_{t=k-\delta+1}^{k+\delta}
    \left\lVert
    \mathbf{c}_{t}-\mathbf{c}_{t-1}
    \right\rVert ,
\end{equation}
where $c_t$ denotes the centroid of the object mask at frame $t$. 

After obtaining local motion saliency scores, MIC modulates the base reliability scores under expression conditions using motion. It employs the motion importance signal $\mathcal{M}$ provided by MSP to control the degree to which saliency scores participate in this modification.The final score for candidate frame $k$ is defined as:
\begin{equation}
    S_{\mathrm{dynamic}}(k)
        =
        S_{\mathrm{base}}(k)
        \cdot
        \left(
        1
        +
        \alpha \cdot  S_{\mathrm{motion}}(k) \cdot \mathcal{M}
        \right),
\end{equation}
where $\alpha$ controls the overall strength of motion calibration. This formulation ensures that motion cues act only as a corrective factor rather than a defining element.

\textbf{Temporal Decision Making.}
The Temporal decision making phase distinguishes decision processes under different expressive contexts through an explicit gating mechanism. We utilize previously extracted motion signals to assign candidate frames to two independent scoring branches. When the expression's semantics do not require motion cues, the model makes decisions using base score. When the expression contains motion semantics, the motion-aware calibration branch is activated to further refine the base score. The final scoring of candidate frames is formally defined as:
\begin{equation}
    S_{\mathrm{final}}(k)
    =
    \begin{cases}
    S_{\mathrm{base}}(k), & \text{if } \mathcal{G}=0, \\[4pt]
    S_{\mathrm{dynamic}}(k), & \text{if } \mathcal{G}=1,
    \end{cases}
\end{equation}
After obtaining the final score for each candidate frame, the keyframe is selected by maximizing the score of candidates:
\begin{equation}
    k^{\ast}
    =
    \arg\max_{k \in \mathcal{K}}
    S_{\mathrm{final}}(k),
\end{equation}
The selected keyframe $k^{\ast}$ is then treated as the initial anchor frame for video object segmentation. The object mask predicted on the anchor frame is propagated forward and backward in time using a bi-directional propagation strategy, producing temporally consistent segmentation masks for all frames in the video sequence.
\section{Experiment}
\begin{table*}
    \caption{Performance comparison with previous methods on Ref-YouTube-VOS, Ref-DAVIS17 and MeViS. The highest result in each column will be marked in \textbf{bold}, and the second-highest result will be \underline{underlined}.}
    \setlength{\tabcolsep}{10.1pt}
        \begin{tabular}{ccccccccccc}
            \toprule
            \multirow{2}{*}{Method} 
            & \multirow{2}{*}{Publication}
            & \multicolumn{3}{c}{Ref-YouTube-VOS} 
            & \multicolumn{3}{c}{Ref-DAVIS17}
            & \multicolumn{3}{c}{MeViS} \\
            \cmidrule(lr){3-5} \cmidrule(lr){6-8} \cmidrule(lr){9-11} 
            &
            & $\mathcal{J}\&\mathcal{F}$ & $\mathcal{J}$ & $\mathcal{F}$
            & $\mathcal{J}\&\mathcal{F}$ & $\mathcal{J}$ & $\mathcal{F}$
            & $\mathcal{J}\&\mathcal{F}$ & $\mathcal{J}$ & $\mathcal{F}$ \\
    
            \midrule
            URVOS \cite{seo2020urvos} & ECCV 2020 
            & 47.2 & 45.3 & 49.2 
            & 51.6 & 47.3 & 56.0 
            & 31.0 & 29.8 & 32.2 \\
            \rowcolor[rgb]{0.871,0.918,0.973}
            ReferFormer \cite{wu2022language} & CVPR 2022 
            & 62.9& 61.3 & 64.6 
            & 61.1 & 58.1 & 64.1 
            & 27.8 & 25.7 & 29.9 \\
            MTTR \cite{botach2022end} & CVPR 2022 
            & 55.3 & 54.0 & 56.6 
            & - & - & - 
            & - & - & - \\
            \rowcolor[rgb]{0.871,0.918,0.973}
            HTML \cite{han2023html} & ICCV 2023
            & 63.4& 61.6 & 65.2 
            & 62.1 & 59.2 & 65.1 
            & - & - & -  \\
            SgMg \cite{miao2023spectrum} & ICCV 2023
            & 65.7& 63.9 & 67.4 
            & 63.3 & 60.6 & 66.0 
            & - & - & - \\
            \rowcolor[rgb]{0.871,0.918,0.973}
            LMPM \cite{ding2023mevis} & ICCV 2023
            & 65.7& 63.9 & 67.4 
            & 63.3 & 60.6 & 66.0 
            & 37.2 & 34.2 & 40.2 \\
            MUTR \cite{mutr} & AAAI 2024
            & 68.4& 66.4 & 70.4 
            & 68.0 & 64.8 & 71.3 
            & - & - & -   \\
            \rowcolor[rgb]{0.871,0.918,0.973}
            DsHmp \cite{DsHmp} & CVPR 2024
            & 67.1 & 65.0 & 69.1 
            & 64.9 & 61.7 & 68.1 
            & 46.4 & 43.0 & 49.8 \\
            DMVS \cite{fang2025decoupled} & CVPR 2025
            & 64.3 & 62.4 & 66.2 
            & 65.2 & 62.2 & 68.2 
            & 48.6 & 44.2 & 52.9 \\
            \rowcolor[rgb]{0.871,0.918,0.973}
            SSA \cite{pan2025semantic} & CVPR 2025
            & 64.3 & 62.2 & 66.4 
            & 67.3 & 64.0 & 70.7 
            & 48.9 & 44.3 & 53.4 \\
            SAMWISE \cite{cuttano2025samwise} & CVPR 2025
            & 69.2 & 67.8 & 70.6 
            & 70.6 & 67.4 & 74.5 
            & 49.5 & 46.6 & 52.4 \\
            \rowcolor[rgb]{0.871,0.918,0.973}
            ReferDINO \cite{liang2025referdino} & ICCV 2025
            & 69.3 & 67.0 & 71.5 
            & 68.9 & 65.1 & 72.9 
            & 49.3 & 44.7 & 53.9 \\
            MPG-SAM 2 \cite{MPG-SAM2} & ICCV 2025
            & \underline{73.9} & \underline{71.7} & \underline{76.1} 
            & 72.4 & \underline{68.8} & 76.0
            & \underline{53.7} & \underline{50.7} & \underline{56.7}\\
            \rowcolor[rgb]{0.871,0.918,0.973}
            TQF\cite{zhang2025mitigating} & ACM MM 2025
            & 69.1 & 67.7 & 70.5 
            & 68.6 & 64.3 & 73.0 
            & 52.8 & 49.4 & 56.1 \\
            FlowRVS \cite{wang2026deforming} & ICLR 2026
            & 69.6 & 67.1 & 72.1 
            & \underline{73.3} & 68.4 & \underline{78.2} 
            & 51.1 & 47.6 & 54.6 \\
            \midrule
            \rowcolor[rgb]{0.871,0.918,0.973}
            \textbf{EMC} & Ours
            & \textbf{74.7} & \textbf{72.6} & \textbf{76.8} 
            & \textbf{77.7} & \textbf{74.5} & \textbf{80.8} 
            & \textbf{54.9} & \textbf{52.2} & \textbf{57.6} \\					
            \bottomrule
        \end{tabular}
    \label{tab:table1}
\end{table*}
\subsection{Experimental Setup}
\subsubsection{Datasets and Metrics}
We evaluate EMC on six public RVOS benchmarks: Ref-YouTube-VOS\cite{seo2020urvos}, Ref-DAVIS17\cite{davis}, MeViS\cite{ding2023mevis}, A2D-Sentences\cite{gavrilyuk2018actor} and JHMDB-Sentences\cite{gavrilyuk2018actor}. Ref-YouTube-VOS contains 3,978 videos with approximately 15K referring expressions. Ref-DAVIS17 includes 90 videos and around 1.5K expressions, while MeViS consists of about 2K videos annotated with 28K expressions. A2D-Sentences contains 3.7K videos with 6.6K expressions, and JHMDB-Sentences includes 928 videos with 928 expressions. For MeViS, we conduct experiments on both valid set and valid$^u$ set, where valid$^u$ is mainly used for offline evaluation during training. For Ref-YouTube-VOS, Ref-DAVIS17, and MeViS, performance is evaluated using region similarity $\mathcal{J}$, contour accuracy $\mathcal{F}$, and their average $\mathcal{J}\&\mathcal{F}$. For A2D-Sentences and JHMDB-Sentences, we employ overall IoU (oIoU) and mean IoU (mIoU) metrics.
\subsubsection{Implementaion Details.}
We adopt the pretrained EVF-SAM\cite{evfsam} with a BEiT\cite{bao2022beit} pretrained backbone for single frame segmentation and confidence estimation, and use SAM2\cite{ravi2025sam} for bi-directional temporal propagation. Image–text alignment scores are computed using Alpha-CLIP\cite{sun2024alpha}. Visual features used for similarity computation are extracted with CLIP\cite{radford2021clip}. Motion-related semantic signals are extracted using Meta-Llama-3-8B-Instruct. For hyperparameter settings, we set $\lambda=0.5, w_t=1, w_\tau=1, w_v=1, \beta=0.5$ and $\delta=2$. The motion calibration strength $\alpha$ and the number of stages $n_{cluster}$ are respectively set to 0.2 and 5 on on Ref-YouTube-VOS and A2D-Sentences, 0.4 and 4 on Ref-DAVIS17 and JHMDB-Sentences, and 0.1 and 10 on MeViS, respectively. Our experiments are conducted on a single V100 GPU with 32GB of memory.
\subsection{Quantitative Results}
\begin{table}
\caption{Performance comparison with previous methods on the valid$^u$ set of MeViS.}
    \setlength{\tabcolsep}{9.5pt}
    \begin{tabular}{ccccc}
        \toprule
        \multirow{2}{*}{Method} 
         & \multirow{2}{*}{Publication}
        & \multicolumn{3}{c}{MeViS (valid$^u$)} \\
        \cmidrule(lr){3-5}
        &
        & $\mathcal{J}\&\mathcal{F}$ & $\mathcal{J}$ & $\mathcal{F}$ \\
        \midrule
         LMPM \cite{ding2023mevis} & ICCV 2023
        & 40.2 & 36.5 & 43.9 \\
        \rowcolor[rgb]{0.871,0.918,0.973}
        DsHmp \cite{DsHmp} & CVPR 2024
        & 55.3 & 51.0 & 60.4\\
        VISA-7B \cite{yan2024visa} & ECCV 2024
        & 51.2 & 48.0 & 54.4\\
        \rowcolor[rgb]{0.871,0.918,0.973}
        DMVS \cite{fang2025decoupled} & CVPR 2025
        & 58.3 & 52.6 & 63.9\\
        SSA \cite{pan2025semantic} & CVPR 2025
        & 57.8 & 52.3 & 63.1\\
        \midrule
        \rowcolor[rgb]{0.871,0.918,0.973}
        \textbf{EMC} & Ours
        & \textbf{62.5} & \textbf{59.6} & \textbf{65.4}\\	
        \bottomrule
        \end{tabular}
\label{tab:validu}
\end{table}
\subsubsection{Results on Ref-Youtube-VOS}
As shown in Table ~\ref{tab:table1}, EMC achieves the best overall performance on Ref-YouTube-VOS\cite{seo2020urvos}. Specifically, EMC attains a $\mathcal{J}\&\mathcal{F}$ score of 74.7, outperforming the previous state-of-the-art MPG-SAM2\cite{MPG-SAM2} by 0.8\% and the sencond ReferDINO\cite{liang2025referdino} by 5.4\%. Improvements are consistently observed on both $\mathcal{J}$ score and $\mathcal{F}$ score, across different evaluation settings, indicating that EMC benefits from more accurate mask localization as well as improved boundary quality. The primary improvement is attributed to STSC, where a more structured and compact temporal space enables the model to locate targets with greater accuracy.
\subsubsection{Results on Ref-DAVIS17}
On the Ref-DAVIS17\cite{davis} benchmark, EMC achieves a $\mathcal{J}\&\mathcal{F}$ score of 77.7, significantly outperforms existing state-of-the-art methods. Compared to the previously best performing FlowRVS\cite{wang2026deforming}, it achieved a 4.4\% improvement, and compared to the second best MPG-SAM2, it achieved an 5.3\% improvement. Ref-DAVIS17 is characterized by high quality annotations and relatively short video sequences, where fine-grained spatial accuracy and temporal consistency are particularly important. Under this setting, selectively activating motion calibration only when action semantics are present allows EMC to avoid unnecessary motion interference for expressions dominated by appearance or spatial cues, leading to more stable predictions overall.
\begin{table}
    \caption{Performance comparison with previous methods on A2D-Sentences and JHMDB-Sentences.}
    \setlength{\tabcolsep}{9pt}
    \begin{tabular}{ccccc}
            \toprule
            \multirow{2}{*}{Method} 
            & \multicolumn{2}{c}{A2D-Sentences} 
            & \multicolumn{2}{c}{JHMDB-Sentences} \\
            \cmidrule(lr){2-3} \cmidrule(lr){4-5}
            & oIoU & mIoU 
            & oIoU & mIoU   \\
            \midrule
            ReferFormer \cite{wu2022language} 
            & 78.6 & 70.3 
            & 73 & 71.8  \\
            \rowcolor[rgb]{0.871,0.918,0.973}
            SgMg \cite{miao2023spectrum} 
            & 79.9 & 72 
            & 73.7 & 72.5  \\
            SOC\cite{luo2023soc} 
            & 80.7 & 72.5 
            & 73.6 & 72.3  \\
            \rowcolor[rgb]{0.871,0.918,0.973}
            WaveCL\cite{chen2025wavecl} 
            & 79.2 & 71.0
            & 73.0 & 71.8  \\
            DAVLS \cite{cao2026dynamic} 
            & 80.9 & 72.4 
            & 73.6 & 72.5  \\
            \midrule
            \rowcolor[rgb]{0.871,0.918,0.973}            
            \textbf{EMC} 
            & \textbf{81.7} & \textbf{72.6} 
            & \textbf{73.8} & \textbf{73.0} \\					
            \bottomrule
        \end{tabular}
    \label{tab:a2d}
\end{table}
\subsubsection{Results on MeViS}
We conduct experiments on both the valid and valid$^u$ sets of the MeViS dataset.
On the MeViS valid$^u$ split, which is a smaller offline evaluation set used during training, EMC achieves the best performance with a $\mathcal{J}\&\mathcal{F}$ score of 62.5. Compared with the previous best method DMVS\cite{fang2025decoupled}, EMC yields a clear improvement of 4.2\%. This improvement is reflected in both $\mathcal{J}$ and $\mathcal{F}$.
On the valid subset of MeViS dataset, EMC further achieves the best results among all comparison methods, with a $\mathcal{J}\&\mathcal{F}$ score of 54.9. Compared to the previous SOTA method MPG-SAM, EMC improved the overall metric by 1.2\%; this improvement was reflected in both $\mathcal{J}$ and $\mathcal{F}$, with a more pronounced increase in $\mathcal{J}$ and a stable gain in $\mathcal{F}$. Furthermore, compared with other beyond the second best approach, EMC maintains a stable performance margin across different evaluation settings. ReferDINO achieves a $\mathcal{J}\&\mathcal{F}$ score of 49.3 on MeViS, while SSA\cite{pan2025semantic} reports 48.9, and EMC outperforms them by approximately 5.6\% and 6.0\%, respectively. These results show that explicitly calibrating motion cues leads to more reliable segmentation under complex motion conditions.
\subsubsection{Results on A2D-Sentences and JHMDB-Sentences}
On the A2D-Sentences\cite{gavrilyuk2018actor} and JHMDB-Sentences\cite{gavrilyuk2018actor} benchmarks, EMC also demonstrates competitive performance, as shown in Table ~\ref{tab:a2d}. On A2D-Sentences, EMC achieves an oIoU of 81.7 and an mIoU of 72.6, outperforming all previous methods. Compared with the strong baseline DAVLS, EMC improves oIoU by 0.8\%. It also achieves the highest mIoU among all methods, slightly surpassing SOC, indicating consistent improvements in both overall overlap and per-instance segmentation quality. On JHMDB-Sentences, EMC achieves the best performance in terms of mIoU, surpassing all existing methods, while maintaining competitive oIoU performance. Compared with the previous best method SgMg, EMC improves mIoU by 0.4\%. Overall, the results on these two datasets further validate the generalization capability of EMC in practice. 
By explicitly calibrating motion influence according to expression semantics, the proposed method consistently improves segmentation quality across datasets with different motion characteristics.
\subsection{Ablation Study}
\subsubsection{Component Analysis}
As shown in Table~\ref{tab:table2}, we conduct a component analysis on MeViS to examine the contribution of each module in EMC, including STSC, MSP, and MIC. When STSC is disabled, the model will employ to uniform temporal sampling for keyframe selection. The baseline configuration without any components (ID 1) achieves a $\mathcal{J}\&\mathcal{F}$ score of 52.6. Enabling STSC alone (ID 2) improves performance to 53.3 $\mathcal{J}\&\mathcal{F}$, indicating that replacing uniform sampling with expression-aware temporal stage construction provides a more effective temporal candidate space. When both MSP and MIC are enabled without STSC (ID 3), the performance further increases to 53.7 $\mathcal{J}\&\mathcal{F}$. This result suggests that explicitly expression-level action modeling and motion calibration can still provide benefits under uniform sampling, but the improvement remains limited. Finally, combining all components (ID 4) yields the best performance, achieving 54.9 $\mathcal{J}\&\mathcal{F}$, with consistent gains on both $\mathcal{J}$ and $\mathcal{F}$. The additional improvement over ID 3 demonstrates that STSC and motion-aware calibration are strongly complementary mechanisms, and structured temporal stages are essential for fully exploiting expression-driven motion modeling on MeViS.
\begin{table}
    \caption{Ablation study of main components of EMC.}
    \setlength{\tabcolsep}{8.5pt}
    \begin{tabular}{ccccccc}
        \toprule
        \multirow{2}{*}{ID} 
        & \multicolumn{3}{c}{Components}
        & \multicolumn{3}{c}{MeViS}\\
        \cmidrule(lr){2-4} \cmidrule(lr){5-7}
        & STSC & MSP & MIC
        & $\mathcal{J}\&\mathcal{F}$ & $\mathcal{J}$ & $\mathcal{F}$ \\   
        \midrule
        1 
        & \xmark & \xmark & \xmark 
        & 52.6 & 49.8 & 55.3 \\
        \rowcolor[rgb]{0.871,0.918,0.973}
        2 
        & \cmark & \xmark & \xmark 
        &53.3 & 50.5 & 56.1 \\
        3 
        & \xmark & \cmark & \cmark 
        &53.7 & 50.9 & 56.5 \\
        \rowcolor[rgb]{0.871,0.918,0.973}
        4 
        & \cmark & \cmark & \cmark 
        &\textbf{54.9} & \textbf{52.2} & \textbf{57.6} \\	
        \bottomrule
    \end{tabular}
    \label{tab:table2}
\end{table}
\subsubsection{Hyperparameter Analysis}
We analyze the influence of the motion calibration strength $\alpha$ on MeViS, as shown in Table~\ref{tab:table3}. When $\alpha$ varies from 0.05 to 0.3, the performance remains stable, with $\mathcal{J}\&\mathcal{F}$ fluctuating within a narrow range. The best result is achieved at $\alpha$=0.1, yielding a $\mathcal{J}\&\mathcal{F}$ score of 54.9. Smaller values of $\alpha$ slightly weaken the effect of motion calibration, while larger values do not provide further improvements and may introduce redundant motion influence. These results indicate that EMC is not highly sensitive to $\alpha$, excessive or insufficient involvement of motion cues can lead to decreased performance, and a moderate calibration strength is sufficient to achieve robust performance.

Figure~\ref{Fig.main4} reports the effect of the number of semantic temporal stages $n_{cluster}$ on Ref-YouTube-VOS, Ref-DAVIS17. On Ref-YouTube-VOS, performance improves as $n_{cluster}$ increases from 3 to 5, reaching the best result at $n_{cluster}$=5, and gradually decreases when more stages are introduced. A similar trend is observed on Ref-DAVIS17, where the optimal performance is obtained at $n_{cluster}$=4. The performance initially increases with the number of semantic temporal stages, then declines. These results suggest that an intermediate number of temporal stages provides a good balance between temporal coverage and stage coherence, while excessive partitioning may lead to redundant candidates in stage construction process.
\begin{table}
\caption{Ablation study of hyperparameter $\alpha$}
    \setlength{\tabcolsep}{21pt}
    \begin{tabular}{cccc}
        \toprule
        \multirow{2}{*}{$\alpha$} & \multicolumn{3}{c}{MeViS} \\
        \cmidrule(lr){2-4}
        & $\mathcal{J}\&\mathcal{F}$ & $\mathcal{J}$ & $\mathcal{F}$ \\
        \midrule
        0.05 & 54.7 & 52.0 & 57.5 \\
        \rowcolor[rgb]{0.871,0.918,0.973}
        \textbf{0.1}  & \textbf{54.9} & \textbf{52.2} & \textbf{57.6} \\
        0.2  & 54.7 & 52.0 & 57.4 \\
        \rowcolor[rgb]{0.871,0.918,0.973}
        0.3  & 54.8 & 52.1 & 57.5 \\
        \bottomrule
        \end{tabular}
\label{tab:table3}
\end{table}
\begin{figure}[t]
\begin{center}
   \includegraphics[width=\linewidth]{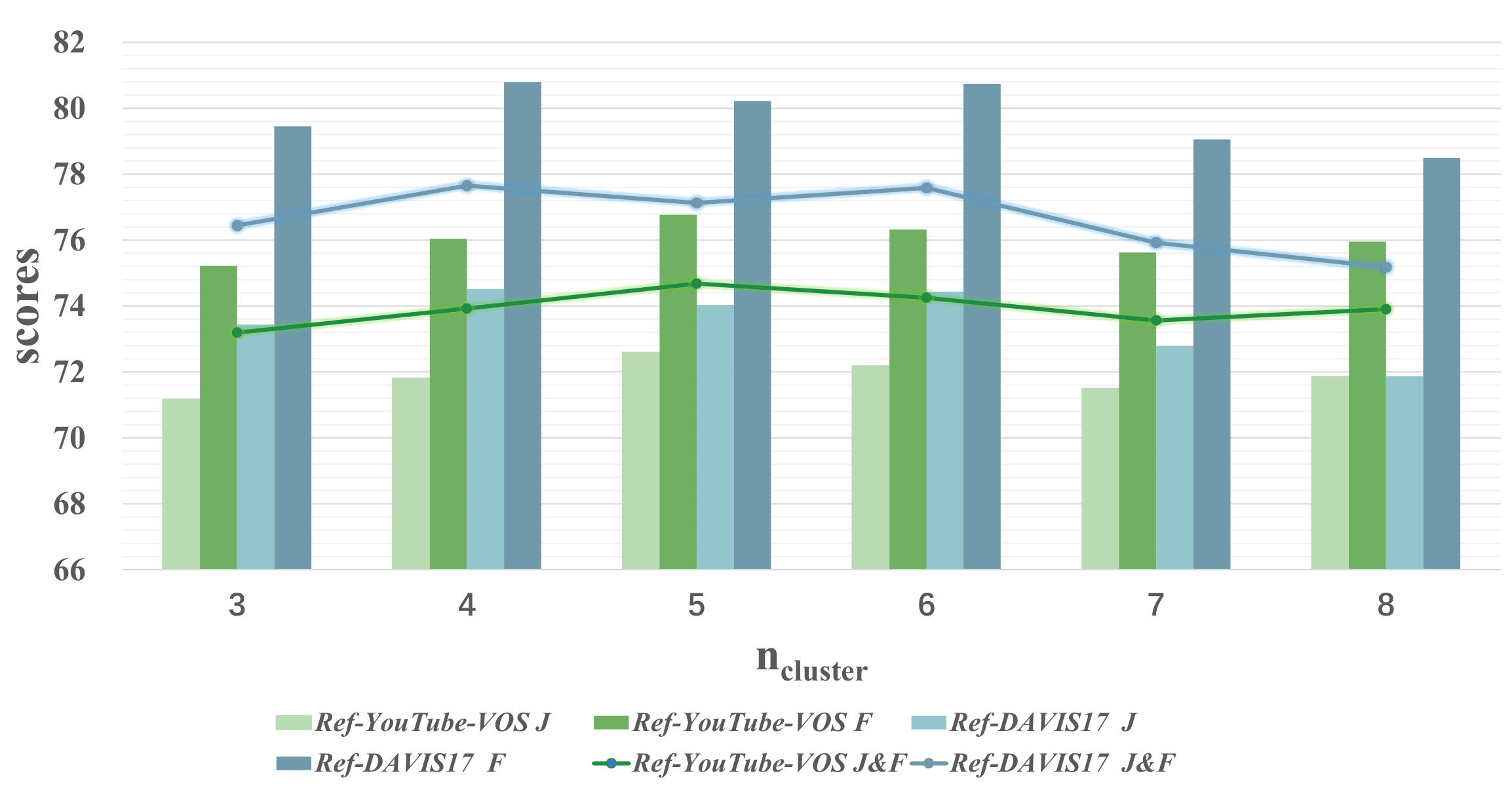}
\end{center}
\vspace{-0.5em}
\caption{Ablation study on number of stages. } 
\label{Fig.main4} 
\end{figure}
\subsubsection{Similarity Components Analysis}
Table~\ref{tab:table5} systematically analyzes the contribution of different similarity components in STSC on MeViS, including visual similarity, text-guided consistency, and temporal proximity similarity. When only visual similarity is used (ID 1), the model achieves a $\mathcal{J}\&\mathcal{F}$ score of 54.5, indicating that appearance consistency provides a reasonable baseline. Using only text-guided similarity (ID 2) leads to lower performance, suggesting that expression signals alone are insufficient. Introducing temporal proximity together with visual similarity (ID 3) yields comparable performance, showing that temporal locality helps stabilize similarity estimation.
Combining cross-modal cues does not always bring improvements. The combination of visual and text similarities (ID 4) results in a performance drop to 53.7, indicating that cross-modal consistency without temporal constraints may introduce noise. When temporal proximity is incorporated with text similarity (ID 5), performance recovers to 54.5. Finally, integrating all three components (ID 6) achieves the best performance of 54.9, demonstrating that these cues jointly provide complementary benefits for constructing reliable semantic temporal stages.
\begin{table}
    \caption{Ablation study of similarity components of STSC.}
    \setlength{\tabcolsep}{8pt}
    \begin{tabular}{ccccccc}
        \toprule
        \multirow{2}{*}{ID} 
        & \multicolumn{3}{c}{Components} 
        & \multicolumn{3}{c}{MeViS}\\
        \cmidrule(lr){2-4} \cmidrule(lr){5-7}
        & Visual & Text & Time
        & $\mathcal{J}\&\mathcal{F}$ & $\mathcal{J}$ & $\mathcal{F}$ \\
        
        \midrule
        1 
        & \cmark & \xmark & \xmark 
        & 54.5 & 51.8 & 57.2 \\
        \rowcolor[rgb]{0.871,0.918,0.973}
        2 
        & \xmark & \cmark & \xmark 
        & 54.4 & 51.7 & 57.1 \\
        3 
        & \cmark & \xmark & \cmark 
        & 54.5 & 51.7 & 57.2 \\
        \rowcolor[rgb]{0.871,0.918,0.973}
        4 
        & \cmark & \cmark & \xmark 
        & 53.7 & 50.9 & 56.4 \\
        5 
        & \xmark & \cmark & \cmark 
        & 54.5 & 51.9 & 57.2 \\
        \rowcolor[rgb]{0.871,0.918,0.973}
        6 
        & \cmark & \cmark & \cmark 
        & \textbf{54.9} & \textbf{52.2} & \textbf{57.6} \\
        			
        \bottomrule
    \end{tabular}
    \label{tab:table5}
\end{table}
\begin{figure}[t]
\begin{center}
\includegraphics[width=\columnwidth]{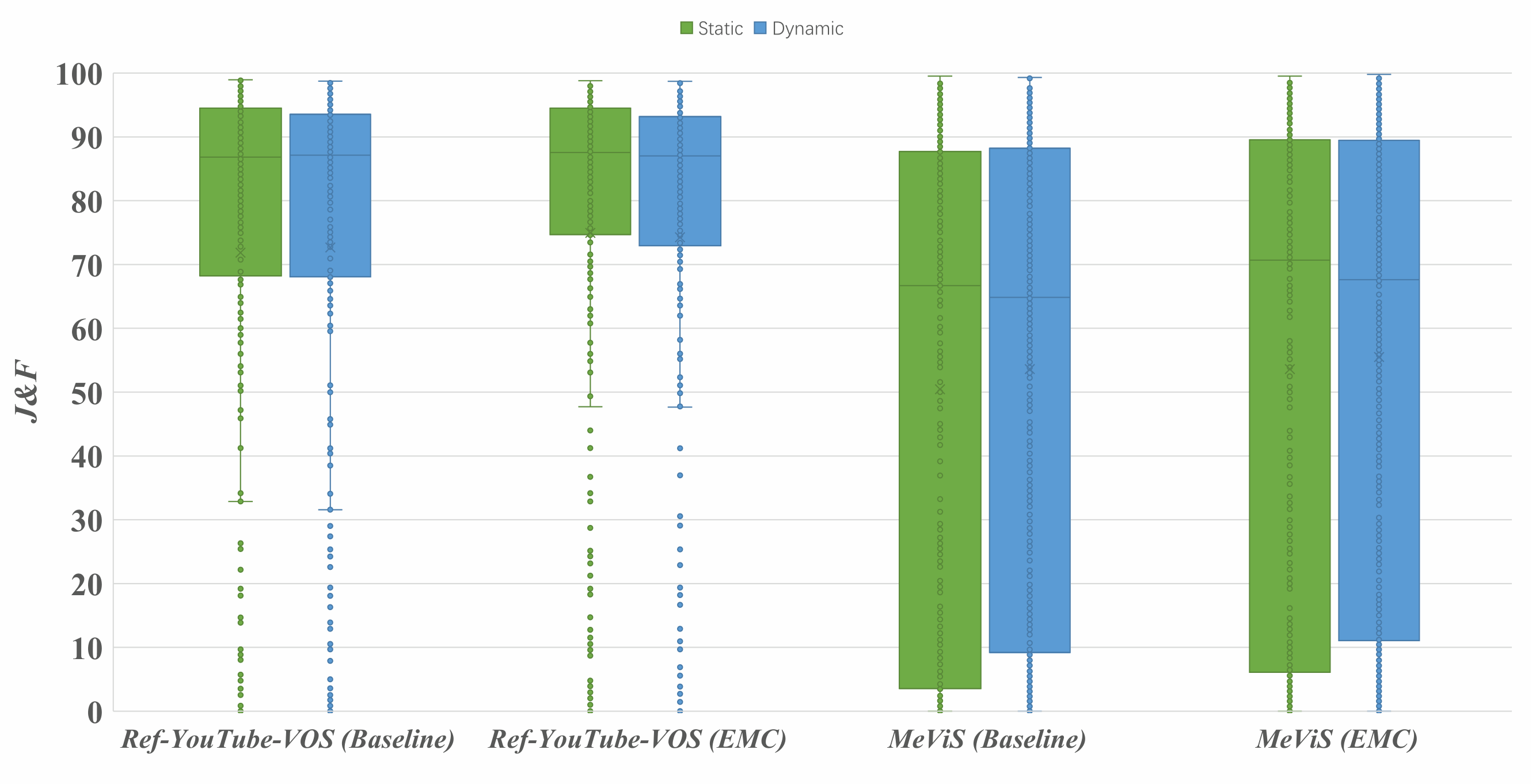} 
\end{center}
\vspace{-0.5em}
\caption{Performance under static and dynamic expressions.} 
\label{Fig.main5} 
\end{figure}
\subsubsection{Behavior under Static and Dynamic Expressions}
We categorize referring expressions into static and dynamic types according to the motion strength inferred by the Motion Signal Processing (MSP) module, and analyze expression-level $\mathcal{J}\&\mathcal{F}$ performance distributions under the two semantic conditions. As shown in Figure~\ref{Fig.main5}, on Ref-YouTube-VOS, the baseline model exhibits noticeable performance variability, particularly for dynamic expressions, with a significant number of low scoring cases. In contrast, after introducing EMC, the performance distributions for both static and dynamic expressions become more concentrated, accompanied by an elevated lower bound and reduced dispersion, indicating improved stability across expressions. On the more challenging MeViS dataset, where dynamic expressions involve complex motion patterns often with intricate temporal dependencies, the baseline method suffers from severe degradation under dynamic conditions, characterized by a pronounced long tail of low performance samples. EMC mitigates this issue by lifting the lower tail of the distribution and narrowing the gap between static and dynamic expressions. These observations demonstrate that explicitly calibrating motion sensitivity based on expression semantics leads to more robust and consistent segmentation behavior across different expression types.
\subsection{Qualitative Results}
\begin{figure}[t]
\begin{center}
\includegraphics[width=\columnwidth]{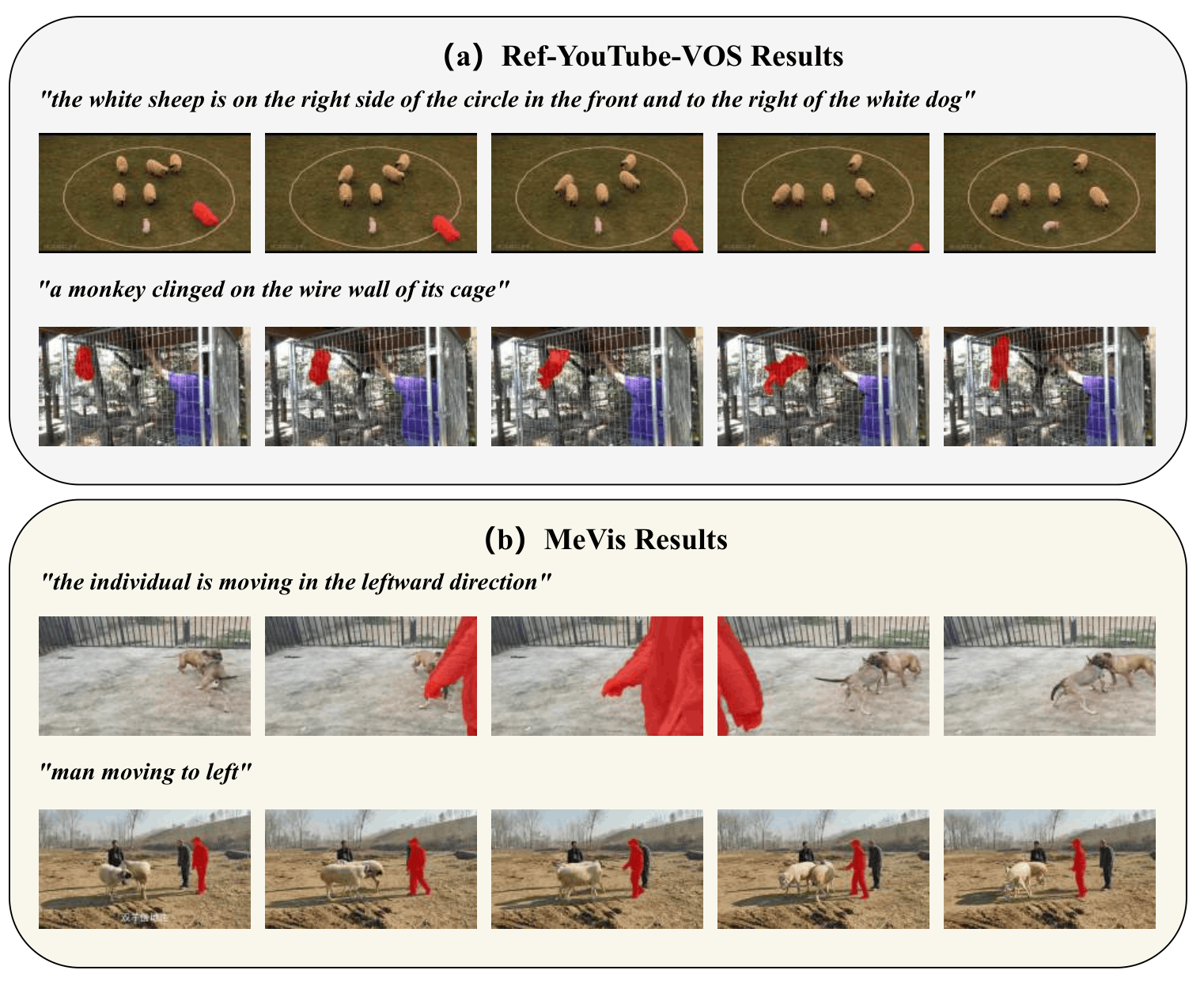} 
\end{center}
\vspace{-0.5em}
\caption{Qualitative results on the Ref-YouTube-VOS (a) and MeVis (b) datasets. Time steps are directed from left to right.} 
\label{Fig.main3} 
\end{figure}

Figure~\ref{Fig.main3} presents qualitative results of our method on the Ref-YouTube-VOS and MeViS datasets. As shown in Figure~\ref{Fig.main3}(a), our method accurately segments dynamic targets on Ref-YouTube-VOS by leveraging motion cues based on expression semantics, demonstrating the effectiveness of motion influence calibration. Specifically, in scenarios where multiple objects share similar appearance, our model is able to precisely distinguish the referred target by jointly modeling spatial position and motion-aware semantics. 
Additionally, even under partial occlusion or cluttered backgrounds, the predicted masks remain temporally stable and consistent across frames, highlighting the robustness of our temporal modeling.
As illustrated in Figure~\ref{Fig.main3}(b), when the target appears only in the middle portion of the video and occupies a relatively small temporal span, our method still successfully localizes and segments the target. 
For example, in motion-centric expressions such as “man moving to left” or “the individual is moving in the leftward direction”,
The model effectively captures fine-grained directional motion cues and focuses on the correct temporal segments. This indicates that STSC can construct expression-related semantic temporal stages, enabling the model to emphasize temporally relevant segments while effectively filtering out irrelevant portions of the video. These qualitative results demonstrate that expression-guided motion calibration improves target localization and temporal consistency effectively.
\section{Conclusion}
In this paper, we propose EMC, a novel expression-driven motion calibration framework. We introduce a motion signal processing module and a motion influence calibration module to perform adaptive forward adjustment of motion cues under diverse expression conditions, with explicit modeling of motion importance based on the expressions. Combined with a semantic temporal stage construction module, this establishes a more precise and effective temporal modeling space for temporal decision making in complex video scenarios. Experimental results on six mainstream datasets, including Ref-YouTube-VOS, Ref-DAVIS17, MeViS(valid/valid$^u$), A2D-Sentences, and JHMDB-Sentences, demonstrate the superior performance and generalization ability of our proposed method across diverse datasets. In the future, we aim to further enhance the robustness and stability of the model across more challenging scenarios.
\section*{Acknowledgments}
This work is supported in part by the National Natural Science Foundation of China (No. 62176172, 61672364); partially by the National Key Research and Development Program of China (No.2018YFA0701700, No. 2018YFA0701701); partially by Basic Research Program of Jiangsu (No. BK20250789).
\bibliographystyle{ACM-Reference-Format}
\bibliography{sample-base}

\clearpage
\appendix
\section{Additional Experimental Analysis}
\subsection{Reliability Analysis of MSP}
To evaluate the reliability of MSP, we randomly sample referring expressions from Ref-DAVIS17, Ref-YouTube-VOS, and MeViS, and manually annotate whether each expression contains motion semantics. The MSP outputs are then compared with human annotations using standard classification metrics.
The quantitative results are summarized in Table~\ref{tab:MSP reliability}. MSP achieves consistently high performance across all datasets, with accuracy ranging from 95\% to 97\%. This demonstrates that MSP can reliably distinguish dynamic expressions from static ones and extract meaningful motion-related signals from natural language. 
A closer look at the metrics reveals that MSP maintains a good balance between precision and recall. In particular, MSP achieves perfect recall on Ref-YouTube-VOS, indicating that motion-related expressions are rarely missed. On Ref-DAVIS17, the slightly lower precision suggests that MSP may occasionally over-detect motion in ambiguous or weak-motion expressions. 
Nevertheless, the overall performance remains stable across datasets.
These results indicate that MSP provides robust and reliable motion signals, which form a solid foundation for subsequent motion calibration and temporal modeling in our framework. 
Furthermore, the reliability of MSP enables consistent motion cues for downstream modules, enhancing overall framework stability.
\subsection{Error Case Analysis of MSP}
To further analyze the limitations of MSP, we examine its prediction errors from both quantitative and qualitative perspectives. 
We first analyze the distribution of error types across different datasets, as summarized in Table~\ref{tab:error distribution}. The results show that over-detection is the dominant error type across all datasets, while weak and implicit motion errors mainly occur in MeViS, which contains more complex and dynamic expressions.
To provide a more intuitive understanding, Table~\ref{tab:error case} presents representative failure cases covering diverse error patterns. As shown in Table~\ref{tab:error case}, the errors of MSP mainly fall into three categories. First, MSP may fail to detect weak or subtle motion, such as “eating”, where motion exists but is not salient or visually prominent. Second, MSP struggles with implicit motion, where dynamic behavior is implied but not explicitly expressed. Third, MSP may over-detect motion when encountering action-related verbs or ambiguous semantics.
These patterns are consistent with the quantitative results, where MSP achieves high recall but relatively lower precision on certain datasets, indicating a tendency toward over-detection. This behavior suggests that MSP prefers to capture potential motion cues rather than risk missing dynamic expressions. Such a design trade-off is beneficial in practice, as missing dynamic cues may lead to more severe errors in subsequent motion-aware temporal decision making and modeling. 
\subsection{Distribution of Motion Strength}
The distribution of extracted motion strength from MSP across different datasets is shown in Figure~\ref{Fig.main1}. Ref-YouTube-VOS and Ref-DAVIS17 exhibit a relatively high proportion of expressions assigned to the lowest motion strength level, indicating that a substantial portion of their referring expressions describe static targets. Meanwhile, both datasets retained a significant proportion of expressions associated with moderate to strong action intensity, reflecting a mixed semantic composition in terms of action intensity.
In contrast, MeViS demonstrates a markedly different distribution pattern, where higher motion strength dominate the overall composition. Specifically, expressions with motion strength values of 0.6 and 0.8 account for the majority, suggesting that dynamic expressions are more prevalent in this dataset.
\begin{table}
    \centering
    \caption{Reliability evaluation of MSP.}
    \vspace{-0.2cm}
    \setlength{\tabcolsep}{2.5pt}
    \begin{tabular}{cccccc}
    \toprule
    Dataset & \#Samples & Acc (\%) & Prec (\%) & Rec (\%) & F1 (\%) \\
    \midrule
    Ref-DAVIS17 & 100 & 96.0 & 90.6 & 96.7 & 93.5 \\
    \rowcolor[rgb]{0.871,0.918,0.973}
    Ref-YouTube-VOS & 150 & 96.7 & 92.3 & 100.0 & 96.0 \\
    MeViS & 200 & 95.5 & 97.9 & 95.9 & 96.9 \\
    \bottomrule
    \end{tabular}
    \label{tab:MSP reliability}
\end{table}
\begin{table}[t]
\centering
\vspace{-0.2cm}
\caption{Error type distribution across datasets.}
\vspace{-0.2cm}
\setlength{\tabcolsep}{4.9pt}
\begin{tabular}{ccccc}
\toprule
Dataset & \makecell{Over-\\detection} & \makecell{Weak\\motion} & \makecell{Implicit /\\Ambiguous} & Total \\
\midrule
Ref-DAVIS17 & 3 & 0 & 1 & 4 \\
\rowcolor[rgb]{0.871,0.918,0.973}
Ref-YouTube-VOS & 5 & 0 & 0 & 5 \\
MeViS & 3 & 3 & 3 & 9 \\
\midrule
\rowcolor[rgb]{0.871,0.918,0.973}
Total & 11 & 3 & 4 & 18 \\
\bottomrule
\end{tabular}
\label{tab:error distribution}
\end{table}
\begin{figure}[t]
\raggedleft
\includegraphics[width=\columnwidth]{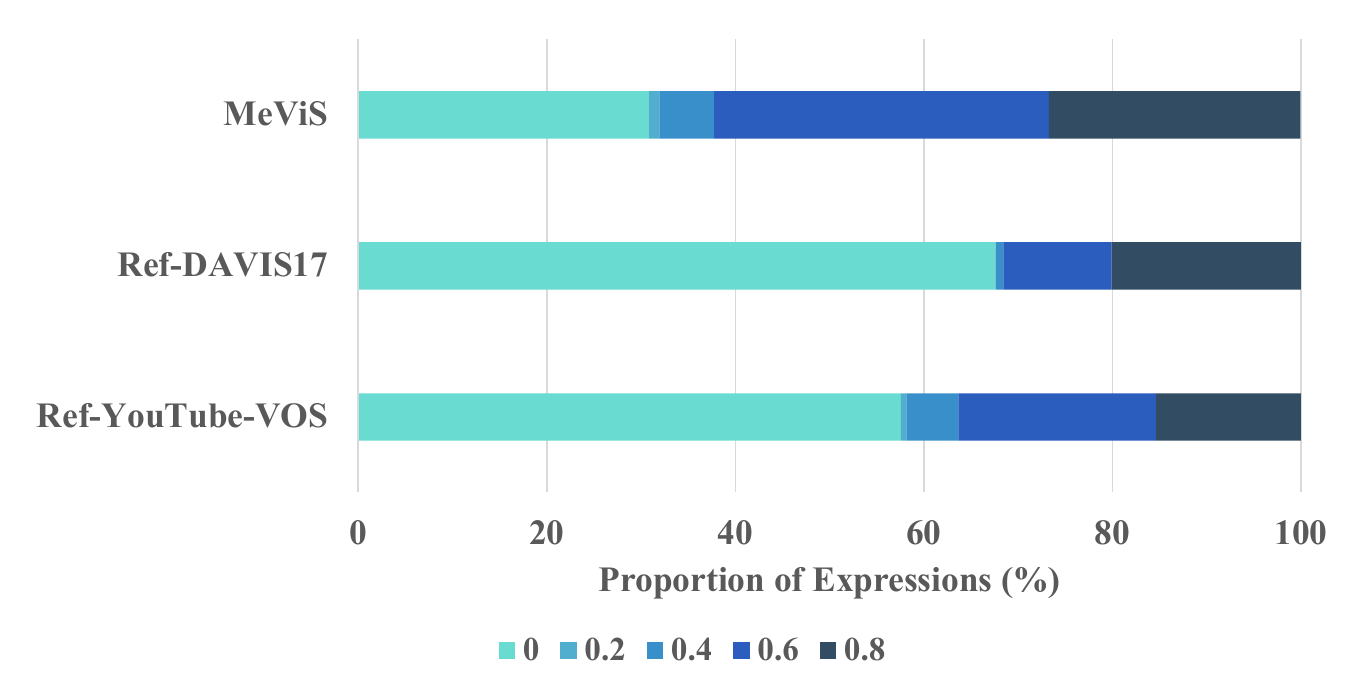} 
\vspace{-0.6cm}
\caption{Distribution of motion strengths across datasets.} 
\vspace{-0.2cm}
\label{Fig.main1} 
\end{figure}
\begin{table*}[t]
\centering
\caption{Representative error cases of MSP.}
\vspace{-0.2cm}
\setlength{\tabcolsep}{0.5pt}
\begin{tabular}{ccccc}
\toprule
Type & Expression & MSP Prediction & Human Annotation & Analysis \\
\midrule
Weak motion & the rabbit that is eating & Static & Dynamic & subtle motion is overlooked \\
\rowcolor[rgb]{0.871,0.918,0.973}
Implicit motion & the last sheep to appear & Static & Dynamic & motion is implied but not explicit \\
Over-detection & a guitar being played by second man & Dynamic & Static & verb triggers false motion detection \\
\rowcolor[rgb]{0.871,0.918,0.973}
Ambiguous action & monkey with a monkey on its back & Dynamic & Static & does not necessarily imply motion \\
Semantic ambiguity & a cow is standing far behind & Dynamic & Static & misinterpreted as motion-related context \\
\bottomrule
\end{tabular}
\label{tab:error case}
\end{table*}
\begin{table}[t]
    \caption{Precision comparison with previous methods on A2D-Sentences and JHMDB-Sentences.}
    \vspace{-0.2cm}
    \centering
    \setlength{\tabcolsep}{4.2pt}

    \begin{tabular}{cccccc}
        \toprule
        \multicolumn{6}{c}{\textbf{A2D-Sentences}} \\
        \midrule
        \multirow{2}{*}{Method} 
         & \multicolumn{5}{c}{Precision} \\
         \cmidrule(lr){2-6}
         & $P@0.5$ & $P@0.6$ & $P@0.7$ & $P@0.8$ & $P@0.9$ \\
        \midrule
        Gavrilyuk et al. \cite{gavrilyuk2018actor}
        & 47.5 & 34.7 & 21.1 & 8.0 & 0.2 \\
        \rowcolor[rgb]{0.871,0.918,0.973}
        MTTR \cite{botach2022end}
        & 75.4 & 71.2 & 63.8 & 48.5 & 16.9 \\
        ReferFormer \cite{wu2022language}
        & 83.1 & 80.4 & 74.1 & 57.9 & 21.2 \\
        \rowcolor[rgb]{0.871,0.918,0.973}
        SOC \cite{luo2023soc}
        & 82.1 & \textbf{82.7} & \underline{76.5} & \underline{60.7} & \underline{25.2} \\
        OnlineRefer \cite{wu2023onlinerefer}
        & 83.1 & 80.2 & 73.4 & 56.8 & 21.7 \\
        \rowcolor[rgb]{0.871,0.918,0.973}
        HTML \cite{han2023html}
        & \underline{84.0} & 81.5 & 75.8 & 59.2 & 22.8 \\
        LAVT \cite{yang2024language}
        & 82.8 & 79.3 & 71.5 & 54.6 & 19.5 \\
        \midrule
        \rowcolor[rgb]{0.871,0.918,0.973}
        \textbf{EMC}
        & \textbf{84.6} & \underline{82.5} & \textbf{77.5} & \textbf{64.3} & \textbf{27.4} \\
        \bottomrule
    \end{tabular}
    \vspace{0.5em}
    \begin{tabular}{cccccc}
        \toprule
        \multicolumn{6}{c}{\textbf{JHMDB-Sentences}} \\
        \midrule
        \multirow{2}{*}{Method} 
         & \multicolumn{5}{c}{Precision} \\
         \cmidrule(lr){2-6}
         & $P@0.5$ & $P@0.6$ & $P@0.7$ & $P@0.8$ & $P@0.9$ \\
        \midrule
        Gavrilyuk et al. \cite{gavrilyuk2018actor}
        & 69.9 & 46.0 & 17.3 & 1.4 & 0.0 \\
        \rowcolor[rgb]{0.871,0.918,0.973}
        MTTR \cite{botach2022end}
        & 93.9 & 85.2 & 61.6 & 16.6 & 0.1 \\
        ReferFormer \cite{wu2022language}
        & 96.2 & 90.2 & 70.2 & 21.0 & 0.3 \\
        \rowcolor[rgb]{0.871,0.918,0.973}
        SOC \cite{luo2023soc}
        & \underline{96.9} & \underline{91.4} & \underline{71.1} & 21.3 & 0.1 \\
        OnlineRefer \cite{wu2023onlinerefer}
        & 96.1 & 90.4 & 71.0 & \underline{21.9} & \underline{0.2} \\
        \midrule
        \rowcolor[rgb]{0.871,0.918,0.973}
        \textbf{EMC}
        & \textbf{97.7} & \textbf{92.3} & \textbf{72.3} & \textbf{23.3} & \textbf{0.3} \\
        \bottomrule
    \end{tabular}

    \label{tab:a2d_and_jhmdb}
\end{table}
\subsection{Additional Precision Comparisons}
As shown in Table~\ref{tab:a2d_and_jhmdb}, we provide additional precision comparisons with previous methods on A2D-Sentences and JHMDB-Sentences under multiple IoU thresholds to evaluate the localization accuracy of different methods under varying levels of strictness. On A2D-Sentences, our method achieves the best performance across most thresholds, especially under higher IoU thresholds. EMC surpasses previous methods with clear margins at $P@0.7$, $P@0.8$, and $P@0.9$, demonstrating its superior capability in precise object localization. Notably, the improvement becomes more pronounced as the IoU threshold increases, indicating that EMC is more effective in capturing fine-grained spatial details and producing accurate segmentation boundaries.
On JHMDB-Sentences, EMC consistently outperforms existing methods across all thresholds. The performance gains are particularly evident at stricter thresholds, further validating the robustness of our approach in scenarios requiring high localization precision.
Overall, these results demonstrate that EMC not only improves overall precision but also shows stronger performance under high-IoU conditions, highlighting its advantage in accurate and robust referring video object segmentation.
\subsection{Visualization of Semantic Temporal Stages}
To qualitatively analyze the behavior of the Semantic Temporal Stage Construction (STSC) module, we visualize the constructed temporal stages on a representative example from the Ref-YouTube-VOS dataset. As shown in the Figure~\ref{Fig.main2}, STSC decomposes the video into multiple semantically coherent temporal stages according to the motion semantics implied in the referring expression. For the expression “a white parachute is opened in the sky and flying downward”, the video is automatically partitioned into five consecutive stages, where frames within each stage exhibit high visual and semantic consistency, while different stages correspond to distinct phases of the downward motion.This structured decomposition reflects the gradual evolution of motion dynamics over time. The red bounding boxes indicate the representative key frames selected for each stage, which effectively summarize the dominant visual and motion characteristics of each stage and provide reliable anchors for subsequent motion-aware calibration.
\begin{figure}[t]
\raggedleft
\includegraphics[width=\columnwidth]{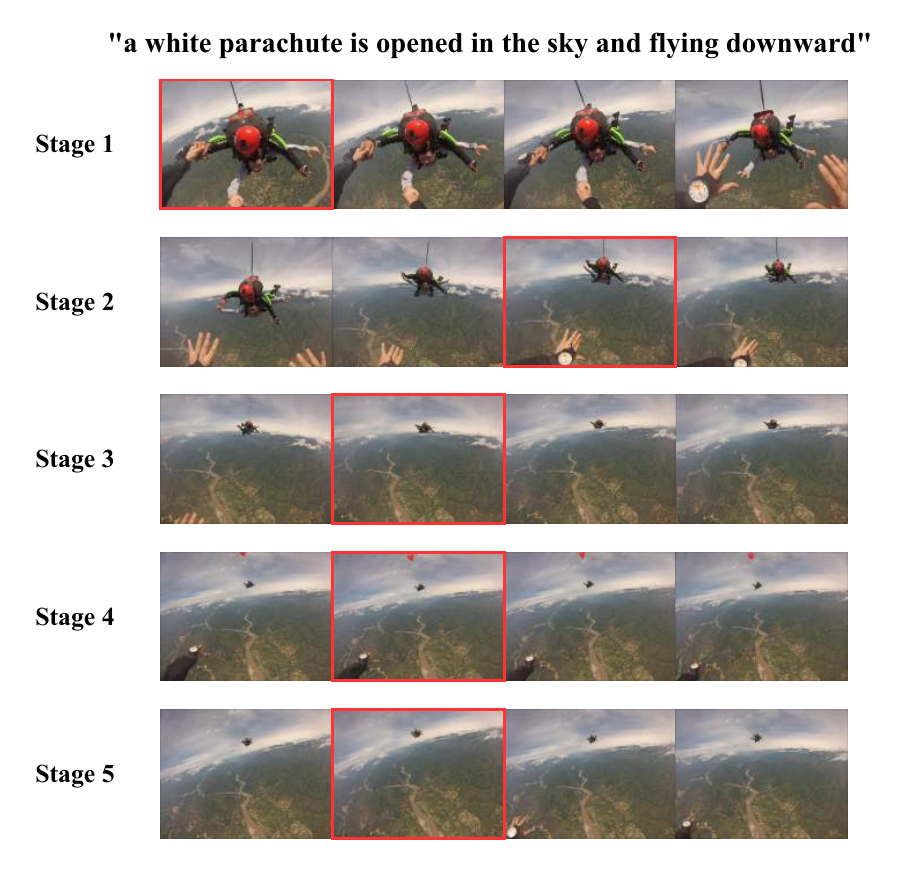} 
\caption{Visualization of Semantic Temporal Stages.} 
\vspace{-0.2cm}
\label{Fig.main2} 
\end{figure}
\subsection{Extra Ablation Study}
\subsubsection{Semantic Temporal Stages on MeViS}
We further conduct an ablation study on MeViS to analyze the effect of the number of semantic temporal stages constructed by STSC. Figure~\ref{Fig.main4} illustrates the performance variation as the number of temporal stages increases. As shown, segmentation performance exhibits a clear upward trend when increasing the number of semantic temporal stages, indicating that finer temporal partitioning enables better alignment with complex motion evolution on MeViS. The performance reaches its peak at a moderate stage number, after which further increasing the number of stages leads to performance saturation and slight degradation. This behavior suggests that while richer temporal decomposition is beneficial, excessively fine partitioning may weaken semantic coherence and introduce redundancy. These results clearly and empirically demonstrate that properly selecting the number of stages is critical for achieving a balance between temporal granularity and semantic consistency on MeViS.
\subsubsection{Motion calibration strength on Ref-DAVIS17}
As shown in table~\ref{tab:table1}, the relatively slow performance improvement observed when increasing the motion calibration strength is closely related to the characteristics of Ref-DAVIS17, where many videos exhibit static or weakly dynamic motion patterns. As the calibration strength increases from 0.1 to 0.4, the $\mathcal{J}\&\mathcal{F}$ score shows a smooth but limited gain, rising from 77.56 to 77.65, with consistent trends in both $\mathcal{J}$ and $\mathcal{F}$. Specifically, the incremental improvements remain marginal across all intervals, indicating a diminishing return as the calibration strength increases. This indicates that while motion calibration contributes positively, moderate calibration is generally sufficient for Ref-DAVIS17 due to its predominantly stable motion content. Moreover, excessive calibration may overemphasize motion cues, yielding limited additional benefit under such scenarios.
\subsubsection{Sensitivity Analysis of $\beta$}
We further investigate the sensitivity of the hyperparameter $\beta$ on Ref-YouTube-VOS, Ref-DAVIS17, and MeViS. As shown in Table~\ref{tab:beta}, the performance remains relatively stable when $\beta$ varies from 0.25 to 1.0, demonstrating that EMC is not sensitive to the specific choice of $\beta$. In particular, $\beta=0.5$ consistently achieves the best performance across all three datasets. In contrast, setting $\beta=0$ leads to a noticeable performance drop, indicating the importance of motion calibration in temporal decision making. Based on these results, we set $\beta=0.5$ as the default value in our experiments.
\begin{figure}[t]
\raggedleft
\includegraphics[width=\columnwidth]{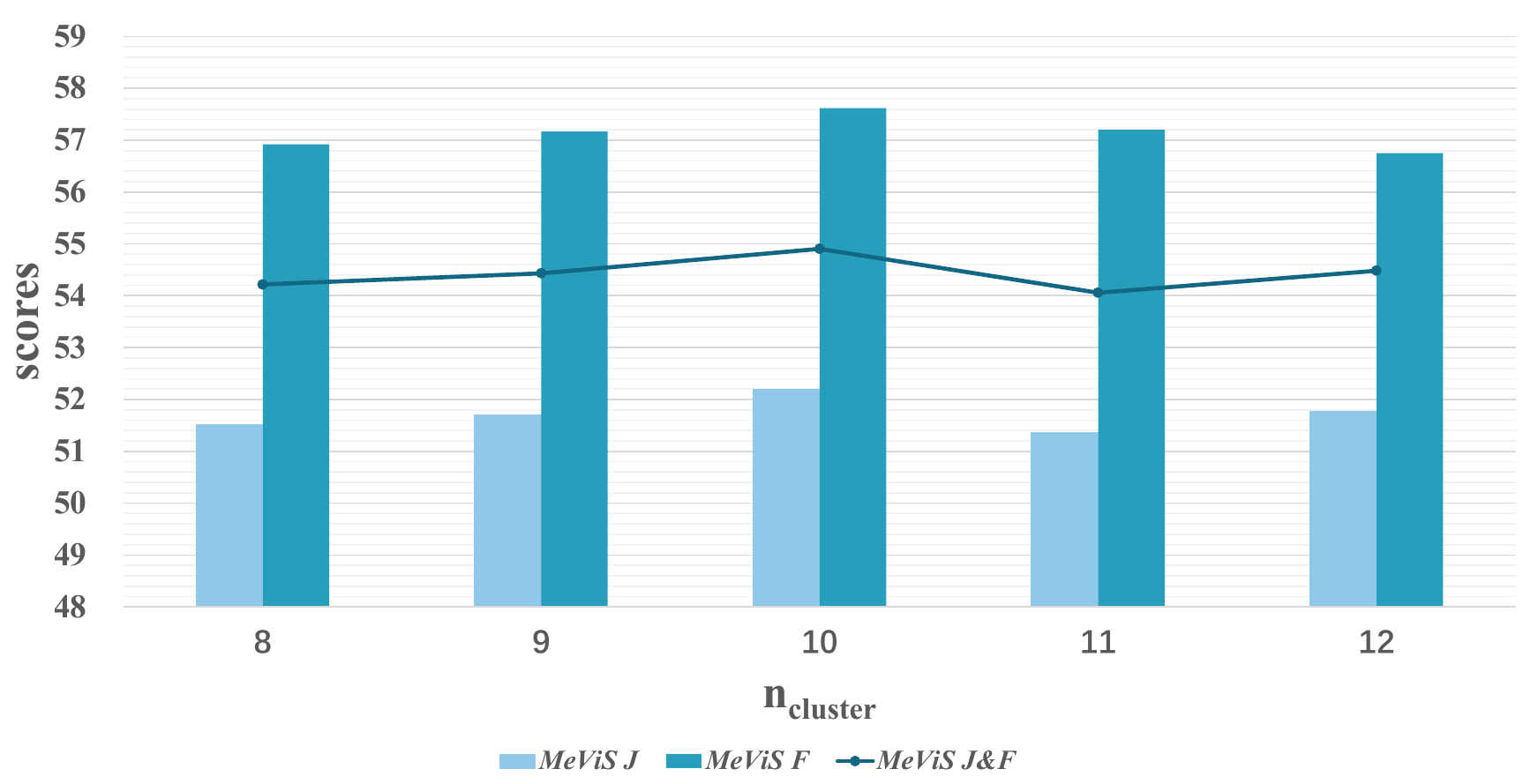} 
\caption{Ablation study on number of stages on MeViS.} 
\vspace{-0.2cm}
\label{Fig.main4} 
\end{figure}
\begin{table}
    \centering
    \caption{Ablation study of motion calibration strength on Ref-DAVIS17.}
    \vspace{-0.2cm}
     \setlength{\tabcolsep}{21pt}
        \begin{tabular}{cccc}
        \toprule
        \multirow{2}{*}{$\alpha$} & \multicolumn{3}{c}{Ref-DAVIS17} \\
        \cmidrule(lr){2-4}
        & $\mathcal{J}\&\mathcal{F}$ & $\mathcal{J}$ & $\mathcal{F}$ \\
        \midrule
        0.1 & 77.56 & 74.41 & 80.71 \\
        \rowcolor[rgb]{0.871,0.918,0.973}
        0.2  & 77.57 & 74.42 & 80.72 \\
        0.3  & 77.63 & 74.49 & 80.77 \\
        \rowcolor[rgb]{0.871,0.918,0.973}
        0.4  & \textbf{77.65} & \textbf{74.51} & \textbf{80.79} \\
        \bottomrule
        \end{tabular}
\label{tab:table1}
\end{table}
\subsection{Inference Efficiency}
We evaluate the inference efficiency of EMC in terms of inference speed, runtime, and memory consumption. As shown in Table~\ref{tab:efficiency}, MSP and STSC achieve 16.81 and 20.11 FPS, respectively, indicating that both preprocessing stages can be efficiently performed. Since their outputs are independent of the motion calibration process, they can be precomputed and cached before inference. With the outputs of MSP and STSC precomputed, the remaining EMC inference process achieves 7.34 FPS, with a runtime of 6.73 seconds per pair and 10.0 GB of memory usage. These results demonstrate that EMC maintains practical inference efficiency while enabling expression-driven motion calibration for temporal decision making.

\section{Inference Details in MSP}
\begin{figure*}[t]
 \centering
\includegraphics[width=\textwidth]{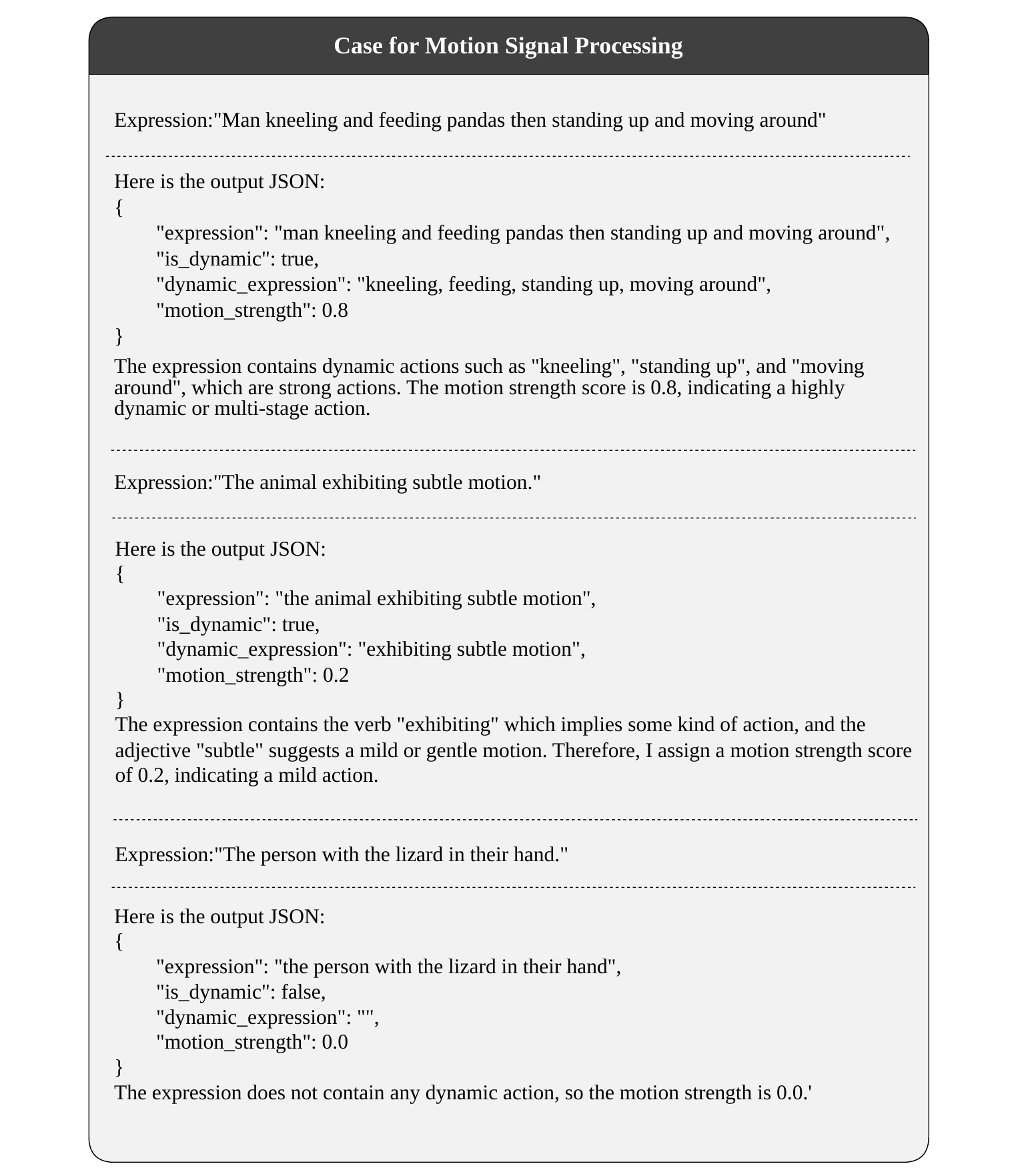} 
\caption{Cases for Motion Signal Processing.} 
\label{Fig.case} 
\end{figure*}

For motion signal extraction, we perform inference using a frozen large language model, Meta-Llama-3-8B-Instruct, without any task-specific fine-tuning. The model is accessed through a standard text generation interface and executed in half precision to reduce memory consumption during inference. For each referring expression, the prompt described above is provided to the model in a single forward pass, and the generated output is directly parsed to obtain structured motion control signals.
To ensure deterministic and stable predictions across different runs, we disable stochastic decoding strategies during inference and adopt greedy decoding with a fixed maximum token budget. This design avoids randomness introduced by sampling based decoding and guarantees consistent motion signal extraction for identical inputs. The inference process is lightweight and performed independently for each expression, incurring negligible overhead compared to overall video segmentation pipeline. The case and prompt are shown in Figure~\ref{Fig.main5} and ~\ref{Fig.case}.
\section{More Visualizations}
\begin{figure*}[t]
 \centering
\includegraphics[width=\textwidth]{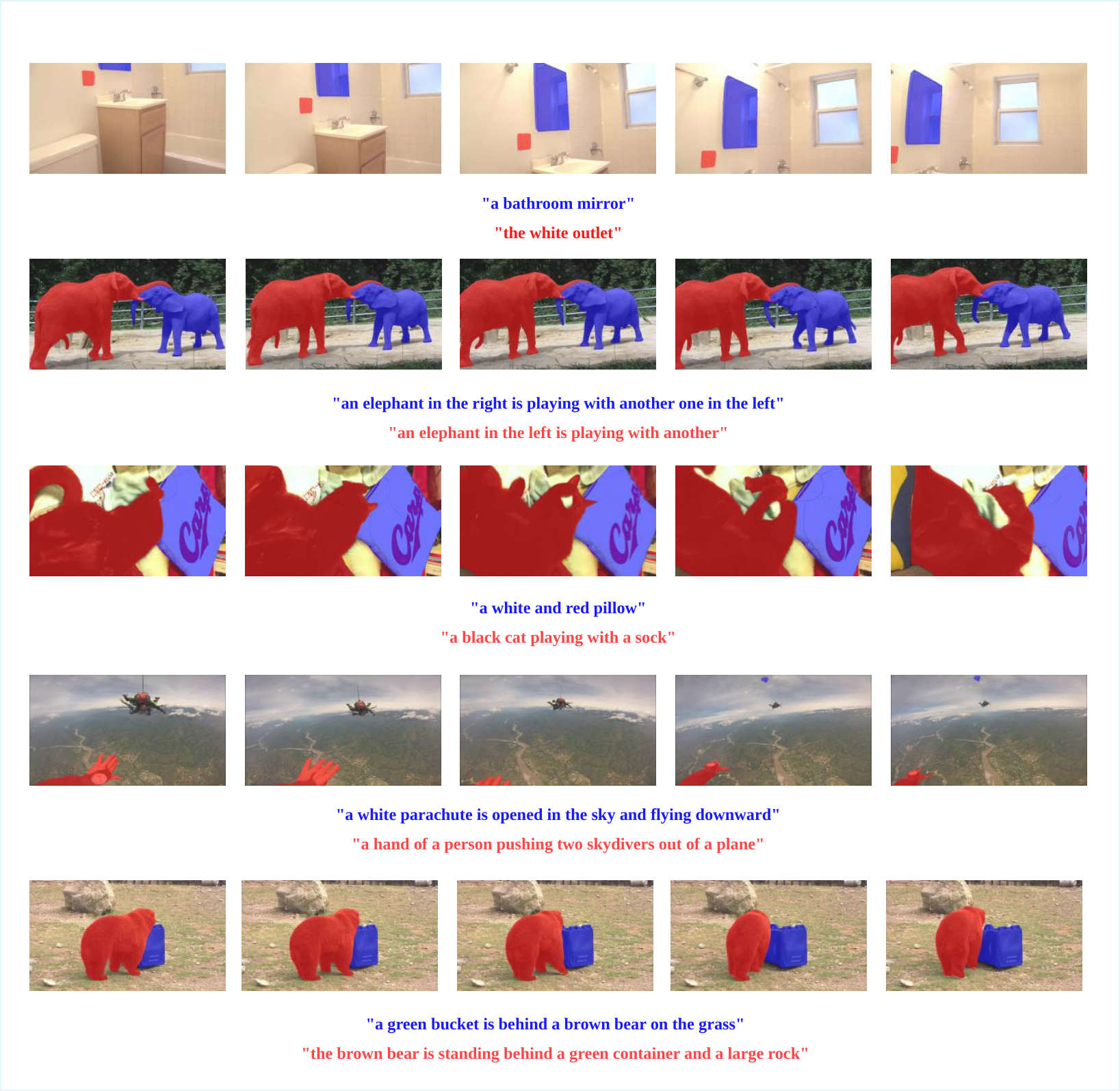} 
\caption{Visualization for multiple text references.} 
\label{Fig.result2} 
\end{figure*}

\begin{figure*}[t]
 \centering
\includegraphics[width=\textwidth]{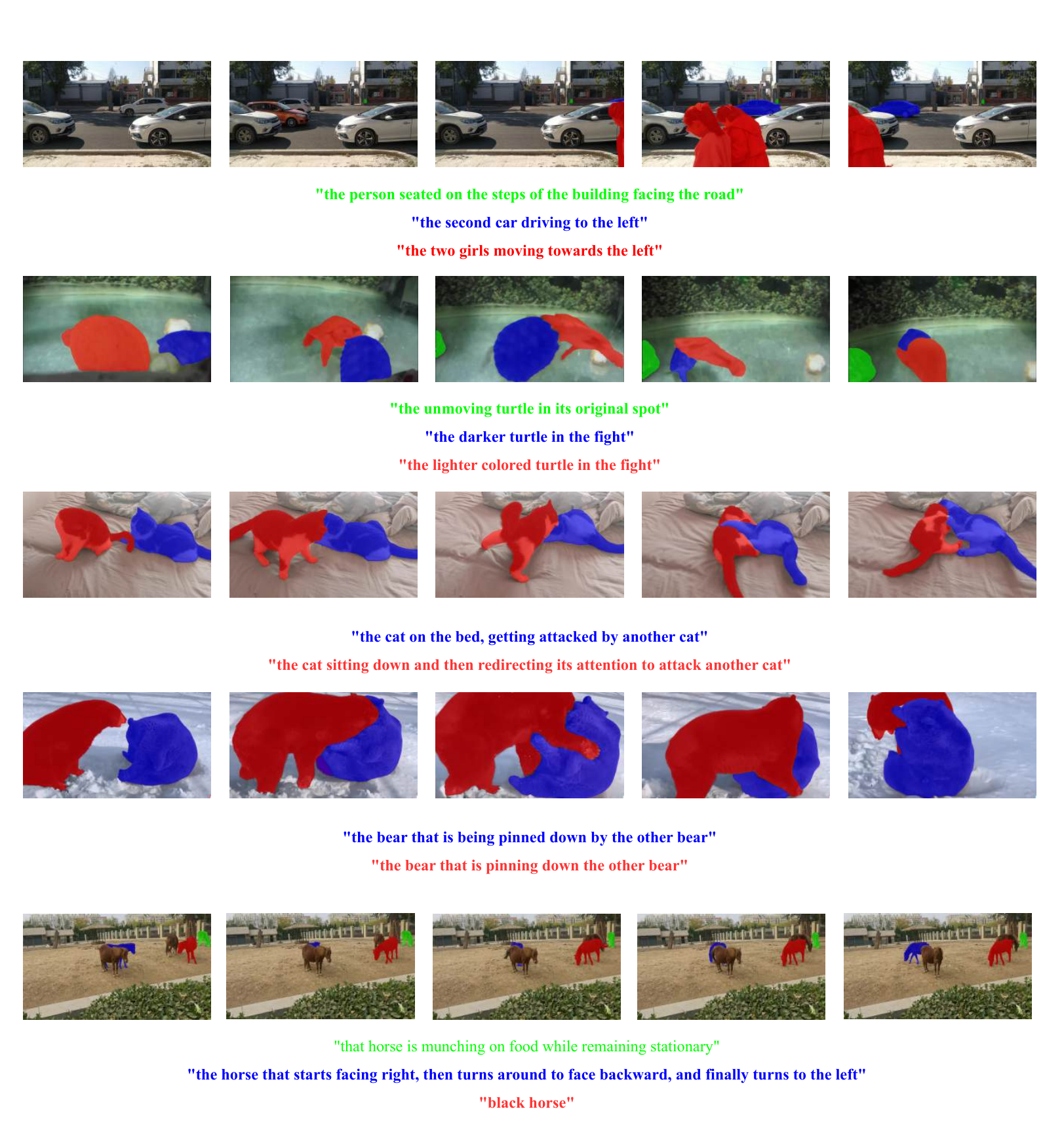} 
\caption{Visualization for multiple text references.} 
\label{Fig.result} 
\end{figure*}

We provide more visualizations of diverse objects in Figure~\ref{Fig.result2} and Figure~\ref{Fig.result} to demonstrate the robustness of our model under diverse referring expressions and object interactions. The examples cover a wide range of scenarios, including static objects, single object motion, and complex multi-object interactions in real scenes with similar appearances or overlapping trajectories. As shown, our method is able to correctly localize and segment the referred target under diverse and challenging practical conditions, even when multiple objects of the same category coexist, or when the expression emphasizes subtle motion differences, role relations, or temporal action changes. These results indicate that the proposed framework can consistently and effectively integrate expression semantics and temporal cues to handle challenging referring video segmentation cases across diverse object types and motion patterns.
\begin{table}[t]
\small
\centering
\caption{Sensitivity analysis of $\beta$.}
\vspace{-0.2cm}
\setlength{\tabcolsep}{4pt}
\begin{tabular}{cccccccccc}
\Xhline{2\arrayrulewidth}
 & \multicolumn{3}{c}{Ref-YouTube} & \multicolumn{3}{c}{Ref-DAVIS17} & \multicolumn{3}{c}{MeViS}\\
 \cmidrule(lr){2-4} \cmidrule(lr){5-7} \cmidrule(lr){8-10}
\multirow{-2}{*}{$\beta$} & \(\mathcal{J} \& \mathcal{F}\) & \(\mathcal{J}\) & \(\mathcal{F}\) & \(\mathcal{J} \& \mathcal{F}\) & \(\mathcal{J}\) & \(\mathcal{F}\) & \(\mathcal{J} \& \mathcal{F}\) & \(\mathcal{J}\) & \(\mathcal{F}\) \\ 
\Xhline{2\arrayrulewidth}
0 & 71.8 & 69.8 & 73.9 & 75.5 & 71.7 & 79.3 & 52.2 & 49.3 & 55.1 \\
\rowcolor[rgb]{0.871,0.918,0.973}
0.25 & 74.4 & 72.4 & 76.3 & 77.5 & 74.3 & 80.7 & 54.8 & 52.1 & 57.5 \\
\textbf{0.5} & \textbf{74.7} & \textbf{72.6} & \textbf{76.8} & \textbf{77.7} & \textbf{74.5} & \textbf{80.8} & \textbf{54.9} & \textbf{52.2} & \textbf{57.6}\\ 
\rowcolor[rgb]{0.871,0.918,0.973}
0.75 & 74.2 & 72.3 & 76.1 & 77.1 & 73.9 & 80.3 & 54.5 & 51.8 & 57.2 \\
1 & 74.1 & 72.2 & 76.0 & 77.1 & 74& 80.2 & 54.4 & 51.7 & 57.1 \\\Xhline{2\arrayrulewidth}
\end{tabular}
\label{tab:beta}
\end{table}
\begin{table}[t]
\centering
\vspace{-0.2cm}
\caption{Inference efficiency of EMC.}
\vspace{-0.2cm}
\setlength{\tabcolsep}{6pt}
\begin{tabular}{cccc}
\Xhline{2\arrayrulewidth}
Setting & FPS (frames/s) & Runtime (s/pair) & Memory (GB) \\
\Xhline{2\arrayrulewidth}
MSP & 16.81 & 1.57 & 15.0 \\
\rowcolor[rgb]{0.871,0.918,0.973}
STSC & 20.11 & 1.31 & 1.3 \\
EMC & 7.34 & 6.73 & 10.0 \\
\Xhline{2\arrayrulewidth}
\end{tabular}
\label{tab:efficiency}
\end{table}
\vspace{-0.2cm}
\begin{figure*}[t]
 \centering
\includegraphics[width=\textwidth]{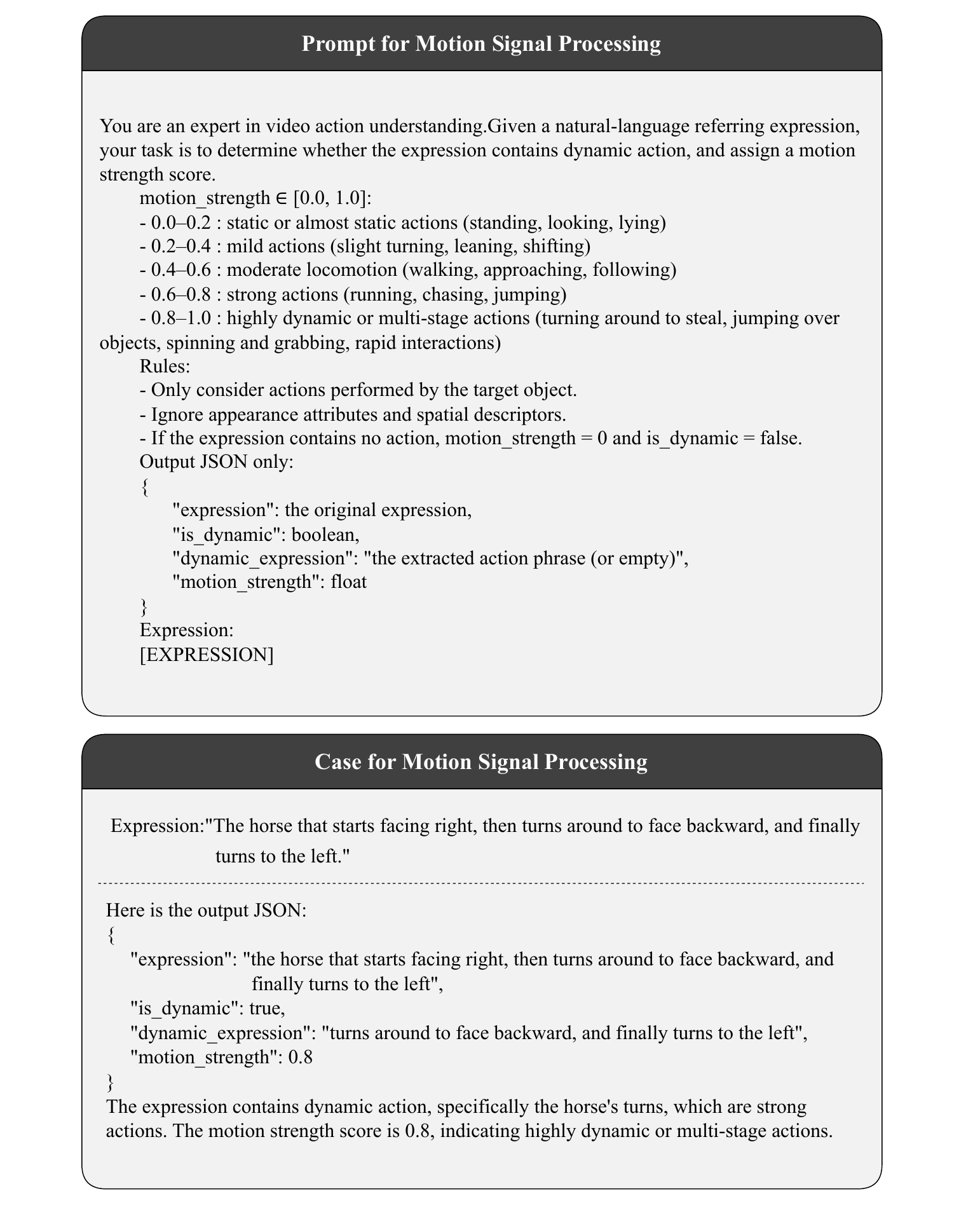} 
\setlength{\abovecaptionskip}{-10pt}
\caption{Prompt and case for Motion Signal Processing.} 
\label{Fig.main5} 
\end{figure*}

\end{document}